\documentclass{article}

\usepackage[nonatbib, preprint]{neurips_data_2024}

\usepackage[utf8]{inputenc} 
\usepackage[T1]{fontenc}    
\usepackage{hyperref}       
\usepackage{url}            
\usepackage{booktabs}       
\usepackage{amsfonts}       
\usepackage{nicefrac}       
\usepackage{microtype}      
\usepackage{xcolor}         

\usepackage{graphicx}
\usepackage{rotating}
\usepackage{multirow}
\usepackage{amsmath}
\usepackage{wrapfig}
\usepackage{nicefrac}

\usepackage{pdflscape}
\usepackage{afterpage}

\newcommand{\figref}[1]{Figure~\ref{#1}}
\newcommand{\tabref}[1]{Table~\ref{#1}}
\newcommand{\secref}[1]{Section~{#1}}
\newcommand{\vtext}[1]{\begin{sideways}#1\end{sideways}}

\title{FindingEmo: An Image Dataset for Emotion Recognition in the Wild}

\author{%
  Laurent Mertens$^{1,2}$  \\
  \And Elahe' Yargholi$^3$ \\
  \And Hans Op de Beeck$^3$ \\
  \AND Jan Van den Stock$^4$ \\
  \And Joost Vennekens$^{1,2,5}$ \\
  \AND \\
  $^1$KU Leuven, De Nayer Campus, Dept. of Computer Science \\
  J.-P. De Nayerlaan 5, 2860 Sint-Katelijne-Waver, Belgium \\
  $^2$ Leuven.AI - KU Leuven Institute for AI, 3000 Leuven, Belgium \\
  $^3$Department of Brain and Cognition, Leuven Brain Institute,\\Faculty of Psychology~\&~Educational Sciences\\
  KU Leuven, 3000 Leuven, Belgium \\
  $^4$Neuropsychiatry, Leuven Brain Institute \\
  KU Leuven, 3000 Leuven, Belgium \\
  $^5$Flanders Make@KU Leuven, 3000 Leuven, Belgium \\
  \texttt{laurent.mertens@kuleuven.be} \\
}

\begin{document}

\maketitle

\begin{abstract}
We introduce FindingEmo, a new image dataset containing annotations for 25k images, specifically tailored to Emotion Recognition. Contrary to existing datasets, it focuses on complex scenes depicting multiple people in various naturalistic, social settings, with images being annotated as a whole, thereby going beyond the traditional focus on faces or single individuals. Annotated dimensions include Valence, Arousal and Emotion label, with annotations gathered using Prolific. Together with the annotations, we release the list of URLs pointing to the original images, as well as all associated source code.
\end{abstract}

\section{Introduction}
\label{s:introduction}
Computer vision has known an explosive growth over the past decade, most notably due to the resurgence of Artificial Neural Networks (ANNs). For many vision-related tasks, computer models have been developed that match or exceed human performance, e.g., image classification \cite{HeKaiming2015DDiR} and mammographic screening \cite{McKinneyScottMayer2020Ieoa}. Many of these tasks, however, are relatively simplistic in nature: detecting the absence or presence of an object, or naming an item in the picture. When it comes to more complex tasks, Artificial Intelligence (AI) still has a long way to go. Affective Computing \cite{AffectiveComputing}, a field that combines disciplines such as computer science and cognitive psychology to study human affect and attempt to make computers understand emotions, is an example of such a complex problem. This paper is concerned in particular with the subtask of Emotion Recognition, i.e.,
building AI models to recognize the emotional state of individuals, in our case from
pictures. This problem has many applications, ranging from psychology \cite{Cowie2001}, to human-
computer interaction \cite{EMANUEL2023104977}, to robotics \cite{Spezialetti2020}. It is, however, complex: in the field of psychology, the concept of what an emotion \emph{is} exactly is heavily debated \cite{barrett1998, barrett2009, Harmon-JonesEddie2017Otio}, resulting in several ways of describing emotions, either by means of continuous dimensions \cite{Russell1980, Mehrabian1996}, or by means of labels, with different competing label classification schemes existing \cite{Ekman1970, Plutchik1980AGP, GenevaEmotionWheel}.

The application of computer vision techniques toward Emotion Recognition has historically largely focused on detecting emotions from human facial expressions, with the problem still being actively investigated \cite{AI_FER_2001, GuoyingZhao2007DTRU, ZhangShiqing2012Rfer, ZhangFeifei2018JPaE, ZhuJunjie2020IFER, Zhang2021, HuangZiYu2023Asoc}. However, the importance of \emph{context} in emotion recognition is increasingly being acknowledged in psychology \cite{Aviezer2012, Kumfor2018}. This led to the release of the computer vision dataset EMOTIC \cite{EMOTIC}, presenting photos of people in natural settings, rather than face-focused close-ups, and leading the way to more complex ANN systems that attempt to combine multiple information streams extracted from these images \cite{EmotiCon, EmoSec, emoreco_mltpl_ctxts}.

\begin{wrapfigure}[20]{R}{7.5cm}
\begin{center}
\includegraphics[width=7.5cm]{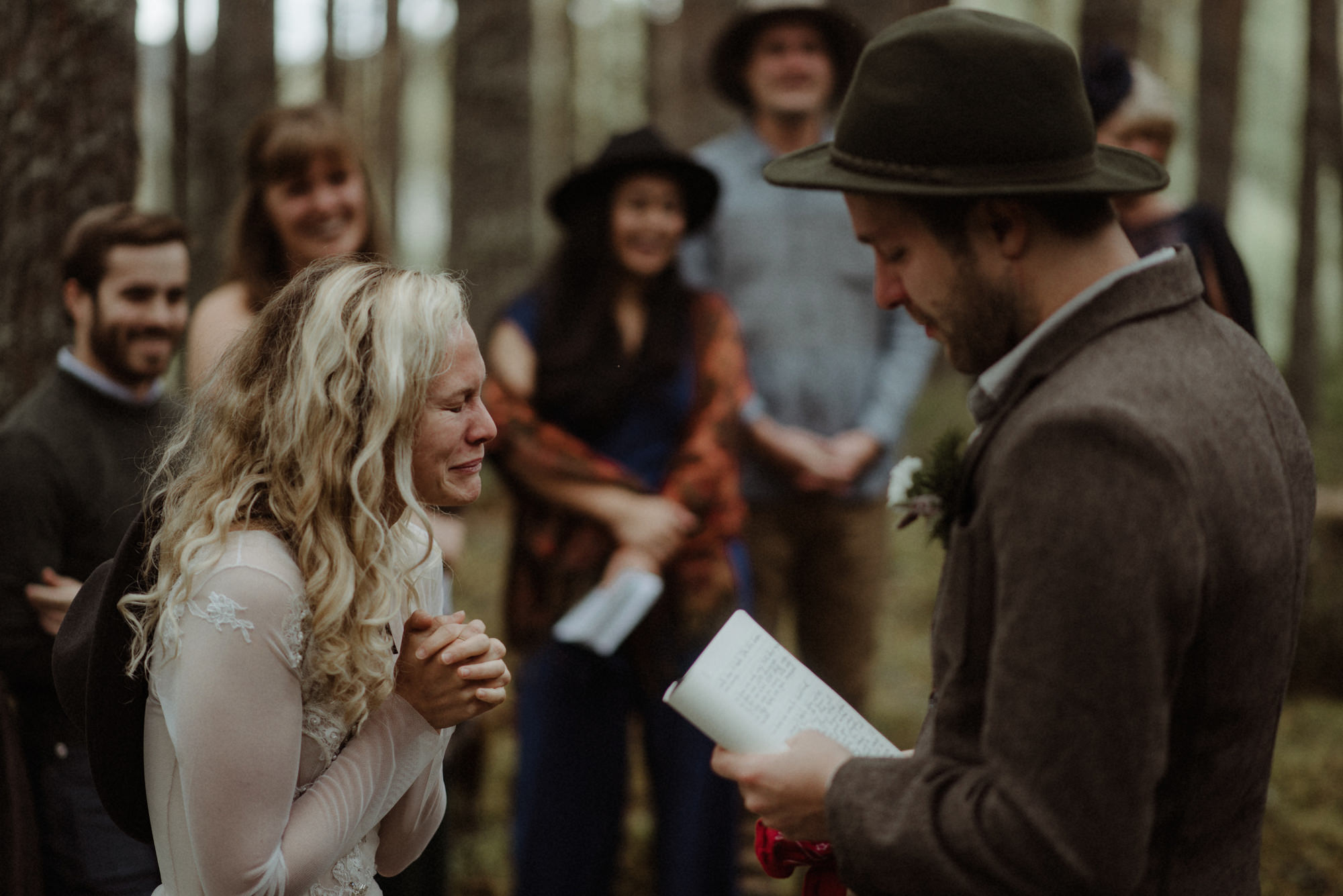}
\caption{An image from the FindingEmo dataset. Photo courtesy The Kitcheners (\url{https://thekitcheners.co.uk/}).}
\label{fig:demo_img}
\end{center}
\end{wrapfigure}
Nevertheless, even these more recent efforts focus on the emotional state of one particular individual within the picture. In this paper, we present the FindingEmo dataset, which is the first to target higher-order social cognition. Each image in the dataset depicts multiple people in a specific social setting, and has been annotated for the \emph{overall} emotional content of the \emph{entire} scene, instead of focusing on a single individual. We hope this data can be used by AI practitioners and psychologists alike to further the understanding of Emotion Recognition, and more broadly, Social Cognition. This is a complex process, consisting of many layers. Consider, e.g., the photograph depicted in \figref{fig:demo_img}. Looking only at the bride's face, one could easily assume she is very sad, or even distressed. Taking also her wedding gown into account, a positive setting is suddenly suggested; perhaps her tears are tears of joy? Only when looking at the full picture does it become clear that the bride is overcome with emotion in a positive way, as conveyed by the setting, the groom reading a prepared text and the clearly supportive bystanders. Thus, full understanding of the bride’s emotional state requires the full scene, including the groom and the solemnly smiling bystanders. This example illustrates how Social Cognition involves detection of relevant elements, extracting relations among these and attributing meaning to construct a coherent whole.

The source code for the scraper and annotation interface used to create the dataset are available from our dedicated repository\footnote{\url{https://gitlab.com/EAVISE/lme/findingemo}}, together with the URLs of the annotated images and their corresponding annotations. To mitigate the issue of broken URLs, we provide multiple URLs for a same image whenever possible, and are continuously expanding the set of images for which multiple URLs are provided (about 10k so far). For copyright reasons, we do not share the images themselves. More information with regard to legal compliance can be found in \S\ref{a:legal_note}.

The data collection process was approved by the KU Leuven Ethics Committee.

The remainder of the paper is structured as follows. In \secref{\ref{s:dataset_desc}} the data collection process and dataset are described in detail. Next, baseline results for emotion classification and valence and arousal regression problems based on popular ImageNet ANN architectures, as well as Visual Transformers CLIP and DINOv2, are presented in \secref{\ref{s:baseline_models}}. We build upon this by investigating the effect of merging the features and predictions of several models in \secref{\ref{s:beyond_baseline}}. Finally, we conclude with a discussion in \secref{\ref{s:discussion}}.

\section{Dataset Description}
\label{s:dataset_desc}
The dataset is split into a publicly released set of annotations for 25,869 unique images, and a privately kept set of 1,525 images\footnote{This set is kept private to allow us to use it as a test set for dedicated workshops organized at a later date.}. Each image depicts multiple people in various, naturalistic, social settings. We follow Emotic \cite{EMOTIC} in creating a training (=our public) set with one annotation per image, and a test (=our private) set with multiple annotations per image. In total, 655 participants---a short description of whom can be found in \S\ref{a:ann_stats}---contributed annotations. In what follows, we list the most important annotation dimensions; for a full list, see \S\ref{a:ann_add_dims}.

\paragraph{Valence and Arousal} We used Russell's continuous Valence and Arousal dimensions \cite{Russell1980}, with integer scales $[-3, -2,\dots, 3]$ for Valence and $[0, 1, \dots, 6]$ for Arousal. Arousal was named ``Intensity'' in our annotation interface, as we felt ``Arousal'' might carry a sexual connotation for some users.

\begin{wrapfigure}[22]{L}{7cm}
\begin{center}
\includegraphics[width=7cm]{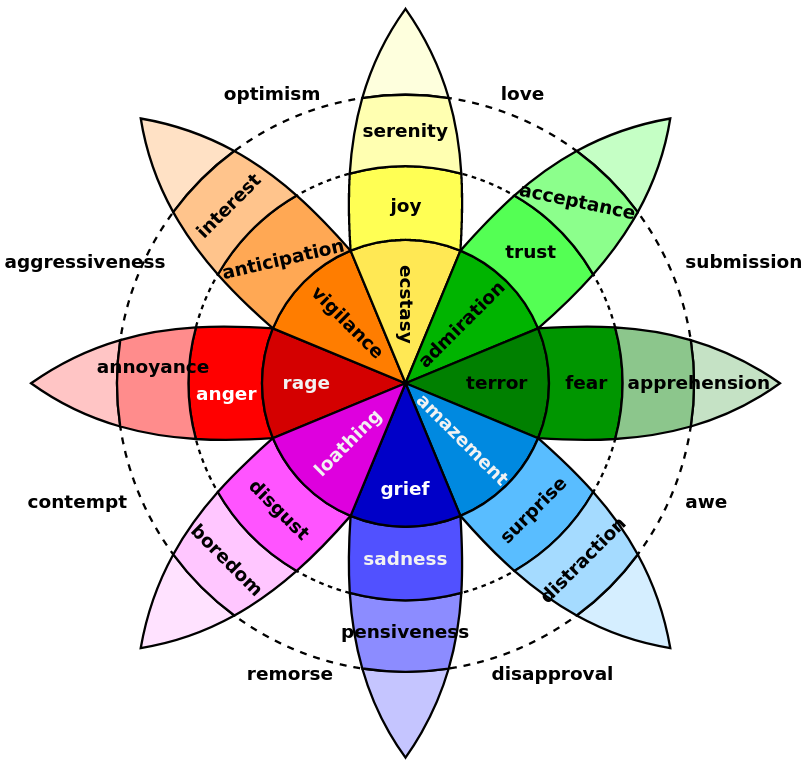}
\caption{Plutchik's Wheel of Emotions.}
\label{fig:pwoe}
\end{center}
\end{wrapfigure}
\paragraph{Emotion} Users had to pick an emotion from Plutchik's discrete Wheel of Emotions (PWoE) \cite{Plutchik1980AGP}, shown in \figref{fig:pwoe}. We opted for this particular emotion classification scheme as it strikes a balance between the more limited and sometimes contested Ekman's 6 \cite{Ekman1970}, and the more expansive, and potentially more confusing, Geneva Emotion Wheel \cite{GenevaEmotionWheel}. It defines 24 primary emotions, grouped into 8 groups of 3, with emotions within a group differing in intensity. It is depicted as a flower with the 24 emotions organized in 8 leaves and 3 concentric rings. Each leaf represents a group of 3, with opposite leaves representing opposite emotions. The rings represent the intensity levels, from most intense at the center to least intense at the outside. An additional advantage of PWoE is that one can easily opt to use all 24 emotions, or instead limit oneself to the 8 groups, allowing some granularity control. We refer to these choices as ``Emo24'' and ``Emo8'' respectively, and refer to the groups as ``emotion leaves''.

\subsection{Positioning Versus Existing Datasets}
\label{ss:positioning}
Although research in automated Emotion Recognition has been gaining in popularity over the years, progress is still hampered by a lack of data. Earlier work tended to focus solely on recognizing emotions from faces. In their recent review paper, \cite{KHARE2024102019} list no less than 21 publicly available datasets of facial images for this purpose, typically annotated with Ekman's 6, potentially extended with a ``neutral'' category, or custom defined emotion categories. Some of the more popular such datasets, like JAFFE \cite{JAFFE} and CK+ \cite{CK+}, make use of a limited number of actors (10 for JAFFE, 123 for CK+) who were instructed to act out a certain emotion, resulting in caricatural emotional expressions.

Publicly available datasets going beyond the face are few in number. First, there is EMOTIC \cite{EMOTIC}, a 23,571 image dataset depicting people in the wild, and with natural expressiveness. An explicit goal of EMOTIC is to take context into account when assessing a person's emotional state. One or more individual subjects are delineated by a bounding box in each picture for a total of 34,320 subjects, each annotated for Valence, Arousal, Dominance and one of 26 custom defined emotion categories.

CAER-S is a dataset of 70,000 stills taken from 79 TV shows. The stills were extracted from 13,201 video clips that were annotated for Ekman's 6 + neutral. Each still contains at least one visible face. The aim of the dataset is to allow augmenting facial emotion recognition with contextual features.

Similar to EMOTIC, there is HECO, a dataset of 9,385 images taken from previously released Human-Object Interaction datasets, films and the internet. Like EMOTIC, 19,781 individual subjects were annotated in the pictures for Valence, Arousal, Dominance, 8 discrete emotion categories comprised of Ekman's 6 + Excitement and Peace, and two novel dimensions, Self-assurance and Catharsis.

\tabref{tb:dataset_comp} groups these dataset descriptions, together with ours, for easy comparison.

\begin{table*}[!htbp]
	\scriptsize
	\centering
	\caption{Comparison of relevant datasets. ``V/A/D'' indicates which of the Valence, Arousal and Dominance dimensions were annotated.}
	\label{tb:dataset_comp}
    \resizebox{1.0\textwidth}{!}{
	\begin{tabular}{@{}l|lp{2.5cm}llp{2.5cm}l@{}}
		Name & Nb. images & Image source & Annotation target & V/A/D & Emotions scheme & Reference \\
		\midrule
		EMOTIC & 23,571 & COCO + Ade20k + internet  & Single person & V/A/D & 26 custom emotion categories & \cite{EMOTIC} \\
		CAER-S & 70,000 & TV Shows & Single person (face visible) & -- & Ekman's 6 + neutral & \cite{CAER-S} \\
		HECO & 9,385 &  HICO-DET + V-COCO + film + internet & Single person & V/A/D & Ekman’s 6 + Excitement and Peace & \cite{emoreco_mltpl_ctxts} \\
		FindingEmo & 25,869 & Internet & Whole image & V/A & Plutchik's Wheel of Emotions & This paper \\
	\end{tabular}
	}
\end{table*}

\subsection{Dataset Creation Process}
\label{ss:dataset_creation_process}
The creation of the dataset was split into two phases. The first phase focused on gathering a large set of images, \emph{prioritizing quantity over quality}. The second phase consisted of collecting the annotations. We present a brief summary of both phases here, and refer to \S\ref{a:dataset_creation} for more details.

\paragraph{Phase 1} Images were gathered using a custom built image scraper that generates random search queries, each consisting of three terms selected from predefined lists of, respectively, emotions, social settings/environments, and age groups of people (e.g., `adults', `seniors', etc.). For each query, the first $N$ results were retrieved, filtered and downloaded. In total 1,041,105 images were collected.

\paragraph{Phase 2} Annotations were gathered using a custom web interface (see \S\ref{a:ann_interface} for a screenshot). Annotators were recruited through the Prolific\footnote{\url{https://www.prolific.com/}} platform, and first required to agree to an Informed Consent clause, followed by detailed instructions (see \S\ref{a:ann_instructions} for a copy). To monitor the process closely, we performed many (51, to be exact) runs, each with a limited number (around 10 to 15) of participants. For each run, the Prolific user selection criteria were the same: fluent English speaker, (self-reported) neurotypical, and a 50/50 split male/female. Candidates were informed of a total expected task duration of 1h, and offered a $\pounds$10 reward. Analysis of the durations (see \S\ref{a:duration_analysis}) show our time estimation to be fair. In total, data collection costs were $\pounds$10k, including fees and taxes.

\subsection{Annotator Grading and Annotator Overlap}
\label{sss:ann_overlap}
To assess the reliability of annotators, we used a set of 5 fixed images, referred to as ``fixed overlap images'', chosen specifically for being unambiguous.\footnote{This amounts to an average of 10\% of the shown images, similar to Emotic \cite{EMOTIC} who ``randomly [inserted] 2 control images in every annotation batch of 20 images''.}  For each image, a default annotation was defined consisting of the ``keep/reject'' choice (4 keeps, 1 reject), Valence (value range), Arousal (value range) and Emotion (emotion leaf). This results in 4 datapoints per image, or 20 datapoints in total. Annotators' submissions for these images were compared to the reference, earning 1/20 point per matching datapoint, resulting in a final ``overlap score'' $s \in [0, 1]$. Users with $s >= 0.8$ were automatically accepted. An alternative score $s_{alt}$ was computed which ignored those overlap images whose reference value was ``keep'', but were annotated as ``reject''. The reason for this is that it quickly became clear that despite the system providing a ``Skip'' option in case users rather not annotate a certain image, some chose to ``reject'' these images instead. Also, one of the ``keep'' images shows a bit of text, which users were instructed to reject. Some users were more strict than others in applying this rule.

We defined a system parameter $p_R$ that controls when overlap images (i.e., images already annotated by others) are shown to users. For each new image request, an overlap image is served with probability $p_R$, starting with the 5 fixed overlap images, in a fixed sequence. Once these are annotated, the system serves other, non-fixed, already annotated images. At first, these were randomly chosen from all annotated images, but this resulted in too many images with only 2 annotations. Hence, we created a process that limits the pool of images to choose from, and attempts to strive for 5 annotations per (non-fixed) overlap image. Using this system, we obtained a dataset with 80.9/19.1 split single label/multi-label annotations. These multi-label images make up the private set. Detailed inter-annotator statistics on this private set are reported in \S\ref{a:inter_ann}, indicating that for 26.2\% of the images, all annotators agreed on the emotion leaf, while for 46.6\% of the images two labels were given. Out of these two-label annotations, 42.8\% refer to adjacent emotion leafs. Annotators agree less on Arousal (average min-max difference of 2.7 $\pm$ 1.4) than on Valence (average min-max difference of 1.8 $\pm$ 1.2). Importantly, average Valence disagreement plateaus close to 2 with increasing number of annotations per image, while a linearely increasing trend is apparent for Arousal.

\subsection{Statistics and Observations}
This section presents statistics for the 8 leaves of PWoE. For the full 24 emotions, see \S\ref{a:extra_data_analysis}.

\begin{wrapfigure}[16]{r}{7cm}
\begin{center}
\centerline{\includegraphics[width=7cm]{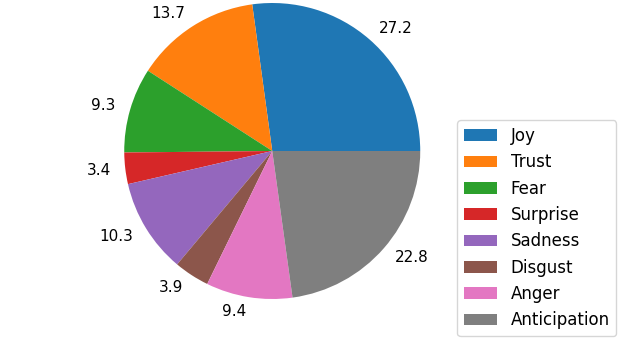}}
\caption{Distribution of Emotion annotations for the public set per Plutchik emotion leaf.}
\label{fig:prolific_emotion_leaf_distribution}
\end{center}
\end{wrapfigure}
\figref{fig:prolific_emotion_leaf_distribution} shows the distribution of annotations per emotion leaf. An imbalance is obvious, with in particular ``joy'' and ``anticipation'' being overrepresented, and ``surprise'' and ``disgust'' heavily underrepresented, despite an added balancing mechanism (see \S\ref{a:sss:gathering_anns}). A similar imbalance is found in popular facial expression datasets, such as FER2013 \cite{FER2013} (only 600 ``disgust'' images versus nearly 5,000 for other Ekman's 6 labels) and AffectNet \cite{AffectNet} (134,915 ``happy'' faces, 25,959 ``sad'' faces, 14,590 ``surprise'' faces, 4,303 ``disgust'' faces). Although EMOTIC \cite{EMOTIC} uses custom emotion labels, making a one-to-one comparison more difficult, it is also heavily skewed towards positive labels (top 3: ``engagement'', ``happiness'' and ``anticipation'';  bottom 3: ``aversion'', ``pain'' and ``embarassement'').
Compared to these other datasets, ours exhibits less imbalance.

\begin{wrapfigure}[20]{l}{7cm}
\begin{center}
\centerline{\includegraphics[width=7cm]{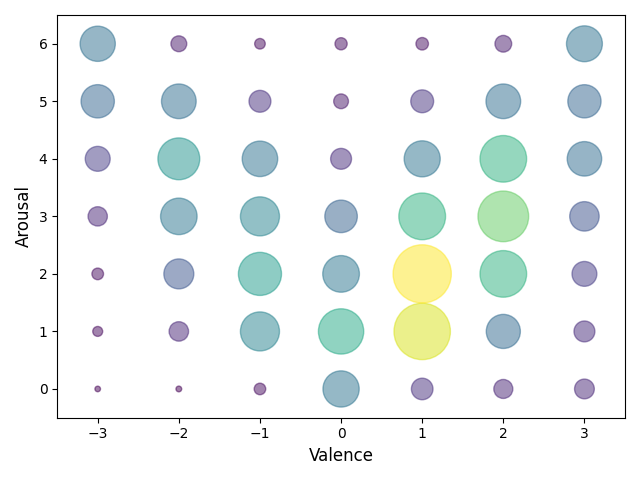}}
\caption{Association between Valence and Arousal values. The bigger the disc, the more often the (Valence, Arousal)-pair appears in the dataset.}
\label{fig:prolific_correlation_valence_arousal}
\end{center}
\end{wrapfigure}In \tabref{tb:emo8_ann_stats}, we group average annotation values for Arousal, Valence and Ambiguity. As expected, perceived ``negative'' emotions (``fear'', ``sadness'', ``disgust'' and ``anger'') have a negative average Valence, with the inverse being true for ``positive'' emotions (``joy'', ``trust''). Somewhat undecided are ``surprise'' and ``anticipation'', which can go either way. The highest Arousal values are reserved for ``anger, ``sadness'' and ``fear''. We hypothesize the unexpectedly high Arousal value for ``sadness'' might be due to naming this dimension ``Intensity'' in our interface; although a grieving person is generally considered to have low arousal, the emotion of sadness itself is felt intensely. Further analysis on the full emotion set reported in \S\ref{a:extra_data_analysis} verifies that also at this more fine-grained level, annotations conform to expectations, with Arousal levels increasing along with the intensity level of the PWoE ring, and Valence levels analogously increasing for ``positive'' and decreasing for ``negative'' emotions.

\begin{table*}[!htbp]
	\centering
	\caption{Average Arousal, Valence and Ambiguity annotation values for the public set, per Plutchik emotion leaf. Format $x^y$: $x$ = average, $y$ = standard deviation.}
	\label{tb:emo8_ann_stats}
    \resizebox{1.0\textwidth}{!}{	
	\begin{tabular}{@{}l|llllllll@{}}
 & Joy & Trust & Fear & Surprise & Sadness & Disgust & Anger & Anticipation \\
 		\midrule
		Nb. & \phantom{$-$}$7026$ & \phantom{$-$}$3549$ & \phantom{$-$}$2401$ & \phantom{$-$}$888$ & \phantom{$-$}$2665$ & \phantom{$-$}$1000$ & \phantom{$-$}$2439$ & \phantom{$-$}$5901$ \\
		Arousal & \phantom{$-$}$2.96^{0.96}$ & \phantom{$-$}$2.57^{1.09}$ & \phantom{$-$}$3.24^{1.24}$ & \phantom{$-$}$2.57^{1.41}$ & \phantom{$-$}$3.42^{1.29}$ & \phantom{$-$}$2.44^{1.23}$ & \phantom{$-$}$3.59^{1.17}$ & \phantom{$-$}$2.46^{1.21}$ \\
		Valence & \phantom{$-$}$1.90^{0.96}$ & \phantom{$-$}$1.41^{1.09}$ & $-1.34^{1.24}$ & \phantom{$-$}$0.48^{1.41}$ & $-1.57^{1.29}$ & $-0.88^{1.23}$ & $-1.58^{1.17}$ & \phantom{$-$}$0.56^{1.21}$ \\
		Ambiguity & \phantom{$-$}$1.58^{1.66}$ & \phantom{$-$}$1.88^{1.64}$ & \phantom{$-$}$2.09^{1.61}$ & \phantom{$-$}$2.39^{1.68}$ & \phantom{$-$}$1.84^{1.66}$ & \phantom{$-$}$2.22^{1.65}$ & \phantom{$-$}$1.99^{1.63}$ & \phantom{$-$}$2.15^{1.61}$ \\
	\end{tabular}
	}
\end{table*}

\figref{fig:prolific_correlation_valence_arousal} shows the association between Arousal and Valence annotations, indicating as expected a collinearity between higher Arousal values and the extremes of the Valence range.

\section{Baseline Model Results}
\label{s:baseline_models}
Baseline results are obtained by applying transfer learning to popular ImageNet-based ANN architectures AlexNet \cite{AlexNet}, VGG16 \cite{VGG}, ResNet 18, 50 and 101 \cite{ResNet} and DenseNet 161 \cite{DenseNet}.\footnote{Our GitLab repository contains all logs used to generate all reported results, and includes additional results for models like VGG19, ResNet34 and DenseNet121, that were in line with other same-architecture models.} For each, we use the default PyTorch implementations and weights, and replace the last layer with a new output layer that matches the chosen task (see below). Only this last layer is trained. We do the same experiment for some of these same architectures trained from scratch on the Places365 dataset \cite{zhou2017places}, using the official PyTorch models. We also consider EmoNet \cite{EmoNet19}, a model for labeling images with one out of 20 custom emotion labels reflecting the emotion elicited in the observer, obtained by applying transfer learning to AlexNet and trained on a private database. In this case, we first process the image with EmoNet, and then send the resulting 20-feature vector through a new linear layer. We use the EmoNet PyTorch port by the main author\footnote{\url{https://gitlab.com/EAVISE/lme/emonet}}. Lastly, we also use Visual Transformer models CLIP \cite{CLIP} (ViT-B/32) and DINOv2 \cite{DINOv2} (ViT-B/14 distilled with registers)\footnote{More specifically the `Pretrained heads for image classification', loaded in PyTorch using \texttt{torch.hub.load(`facebookresearch/dinov2', `dinov2\_vitb14\_reg\_lc')}. We also experimented with the smaller \texttt{vits14} variant, obtaining results typically a few percentage points behind the \texttt{vitb14} model.}, using both models to obtain embeddings for input images, and like with EmoNet, use these as input to a single linear layer.

We distinguish three tasks: \emph{Emo8 classification}, where we predict one of the 8 primary emotions defined by the emotion leaves of PWoE; \emph{Arousal regression}, where we predict the numerical arousal value; \emph{Valence regression}, where we predict the numerical valence value.

\begin{figure}[!htb]
\begin{center}
\centerline{\includegraphics[width=16cm]{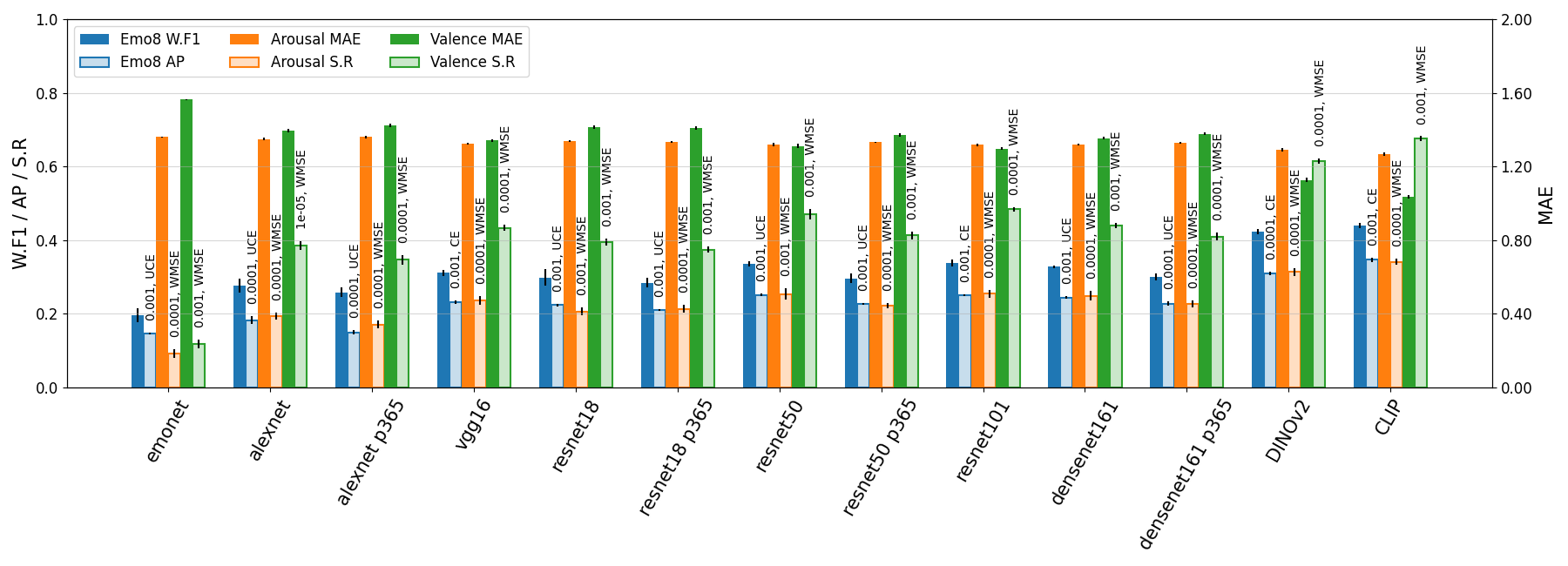}}
\caption{Test data baseline performance on the Emo8 classification and Arousal and Valence regression tasks. Metrics are: Weighted F1 (W.F1) and Average Precision (AP) for classification, and Mean Absolute Error (MAE) and Spearman R correlation coefficient (S.R) for regression. The starting learning rate and loss corresponding to each model are displayed above the training bars. (U)CE = (Unbalanced)CrossEntropyLoss, (W)MSE = (Weighted)MeanSquaredError loss, p365 = original model trained on Places365 dataset.}
\label{fig:barchart_baseline}
\end{center}
\end{figure}

For classification, we apply a softmax to the output of the final layer. Target values for regression problems are reduced to the range $[0, 1]$ using an appropriate linear rescaling. Hence, we apply a sigmoid function to the model output. Network outputs are transformed back to the original problem domain by using the inverse scaling.

Preprocessing for ImageNet models consisted in scaling images to an 800x600 resolution, keeping the original ratio and centering and padding with black borders where necessary, followed by normalization using default ImageNet pixel means and standard deviations. For Places365 and EmoNet models, we followed the preprocessing steps described in the respective papers. For CLIP, we use the default preprocessing chain that comes with the model, and for DINOv2 we use the same preprocessing as for the ImageNet models, but with a rescaling to 798x602.

For each task, and each model, we trained 10 models per starting learning rate $\mbox{lr}_0$ and per loss function $\mathcal{L}$. For classification, we used $\mbox{lr}_0 \in [10^{-1}, 10^{-2}, 10^{-3}, 10^{-4}]$ and $\mathcal{L} \in [\mbox{CrossEntropyLoss}, \mbox{UnbalancedCrossEntropyLoss}]$; for regression we used $\mbox{lr}_0 \in [10^{-3}, 10^{-4}, 10^{-5}, 10^{-6}, 10^{-7}]$ and $\mathcal{L} \in [\mbox{MSELoss}, \mbox{WeightedMSELoss}]$.  UnbalancedCrossEntropyLoss is a novel extension of the traditional CrossEntropyLoss, created to allow giving different weights to different misclassifications. WeightedMSELoss is a natural extension of MSELoss that takes into account class imbalance. Full technical details for both can be found in \S\ref{a:uce_wmse}.

All experiments use the public dataset, Adam loss with default PyTorch parameter values, and the custom lr update rule $\mbox{lr}_{e} = \nicefrac{\mbox{lr}_0}{\sqrt{(e//3) + 1}}$, with $\mbox{lr}_e$ the learning rate at epoch $e$. By virtue of the floor division ($//$), this means we update the learning rate once every 3 epochs. The data was randomly split 80/20 train/test, making sure that each target label was also split according to this same rule.

Reported metrics are: for \emph{classification}, Average Precision (AP)---as computed using the \texttt{scikit-learn} package---and Weighted F1 (W.F1); for \emph{regression}, Mean Average Error (MAE) computed in the original problem domain, and Spearman Rank Correlation (S.R). Training stopped when either the epoch with the best loss (or the best W.F1 score for classification) on the test set lies 6 epochs behind the current epoch, or 250 epochs were reached, with the corresponding best model put forward as the final trained model. Only results for the $(\mbox{lr}_0, \mathcal{L})$-combination yielding the best average Weighted F1 or Mean Average Error performance over the corresponding 10 models are reported.

All our experiments were implemented in Python using PyTorch, and split over an Intel Xeon W-2145 workstation with 32GB RAM and two nVidia GeForce RTX 3060 GPUs with 12GB VRAM, and an Intel i7-12800HX laptop with 32GB RAM and an nVidia GeForce RTX 3060 Laptop GPU with 12GB VRAM. Test results are plotted in \figref{fig:barchart_baseline}, with the graph for train data, and tables containing the numerical results grouped in \S\ref{a:additional_baseline}. In order to speed up training, we buffered model activations whenever possible.\footnote{I.e., we precomputed the output of the frozen part of the model, and stored it on disk for easy reuse.}

Apparent from these results is that these are hard problems. ImageNet-trained models slightly outperform their Places365-trained counterparts. This suggests that the natural object features extracted from the ImageNet dataset are more salient toward emotion recognition than are place-related features. In 9 out of 13 cases, our UnbalancedCrossEntropyLoss has the edge over regular CrossEntropyLoss. Predicting Arousal appears more difficult than predicting Valence, which aligns with lesser annotator agreement for Arousal than Valence, as analyzed in \S\ref{a:inter_ann}. As for the architectures, VGG is a clear winner, with ResNet second. Although twice as large, ResNet101 performs very similar to ResNet50. The larger depth of the DenseNet model does not translate in better performance. A breakdown of model performance per Emo8 class can be found in \S\ref{a:additional_baseline}, showing overall best performance on `Joy' and `Anger'. Worst performance is registered for `Surprise' and `Disgust', which perhaps not surprisingly are also the emotions for which the least annotations are available.

Interestingly, as explored in \S\ref{a:disc_single_anns}, when a model deviates from the target Emo8 annotation there is a strong tendency toward ``nearby'' emotions. Most often this is the adjacent leaf, with more distant leaves increasingly more unlikely. This behavior is reminiscent of the kind of disagreements we find among our human annotators.

\section{Beyond the Baseline}
\label{s:beyond_baseline}
To build upon the baseline established in \S\ref{s:baseline_models}, we built multi-stream models by applying the popular technique of late fusion \cite{EMOTIC, EmotiCon, EmoSec, emoreco_mltpl_ctxts}. This section reports results for Emo8 classification; the analogous discussion for Arousal and Valence regression can be found in \S\ref{a:additional_beyond_baseline}.

\begin{figure}[!htb]
\begin{center}
\centerline{\includegraphics[width=16cm]{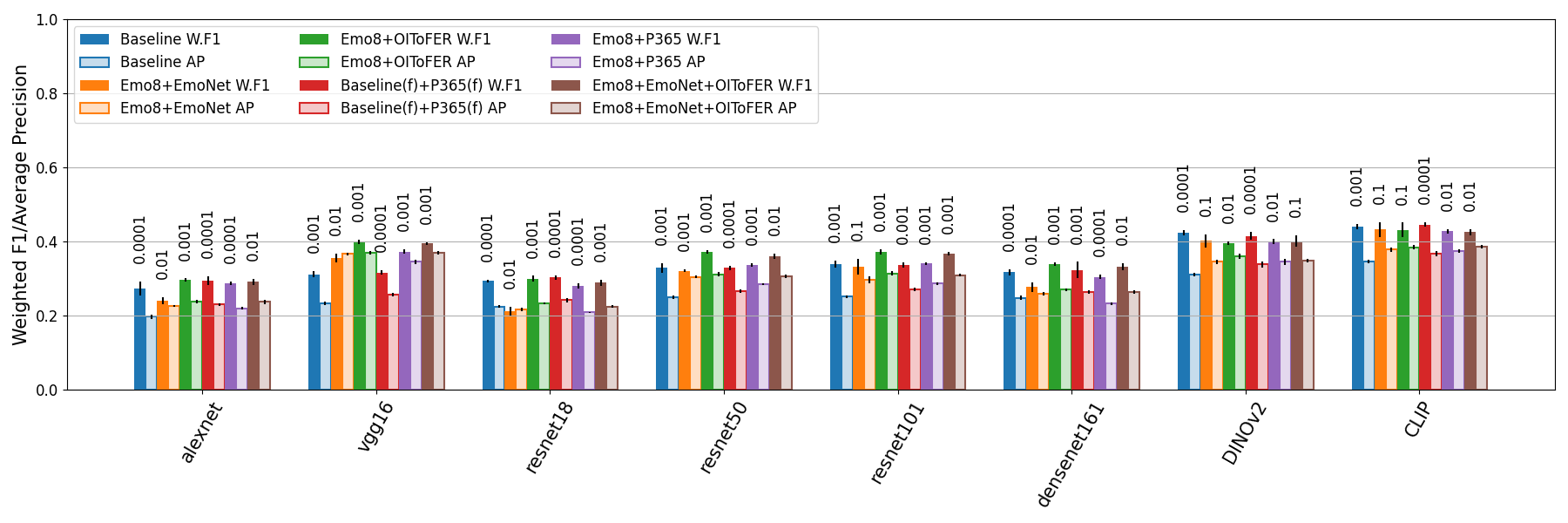}}
\caption{Test data results for extensions beyond the baseline by applying late fusion with Facial Emotion Recognition predictions (OIToFER), EmoNet predictions (EmoNet) and Places365 (P365) predictions or features. For all models, predictions on the dataset (Emo8) are concatenated and sent through a linear layer, except when `(f)' is shown, indicating model features are concatenated. The starting learning rate corresponding to each model is displayed above the training bars.}
\label{fig:beyond_baseline_emo8_test}
\end{center}
\end{figure}

We consider the following streams for combinations:
\emph{Emo8 predictions}: for each considered architecture, we trained an Emo8 model, and took the predictions from this model as an 8-feature vector;
\emph{Baseline features}: we take the model features from the penultimate layer. Vector size depends on the architecture;
\emph{EmoNet predictions}: applying the model gives us a 20-feature vector (see \S\ref{s:baseline_models});
\emph{YoLo v3 trained on Open Images + Facial Emotion Recognition (OIToFER)}: we apply YoLo v3\footnote{We used the Open Images weights available from \url{https://pjreddie.com/darknet/yolo/}.} \cite{Redmon2018YOLOv3AI}, using LightNet \cite{lightnet18}, to each image and extract the detected ``Human face'' regions with probablity $p>0.005$. We then apply the FER2013-trained ResNet18 model by X. Yuan\footnote{\url{https://github.com/LetheSec/Fer2013-Facial-Emotion-Recognition-Pytorch}} to the extracted faces, resulting in a 7-feature vector per face. We generate two 7-feature vectors from this, one containing the vector averages, the other the standard deviations, and concatenate both to obtain a final 14-feature vector;
\emph{Places365 ResNet18 predictions}: applying the ResNet18 model trained on the Places365 dataset gives us a 365-feature vector per image;
\emph{Places365 ResNet18 features}: we take the model activations from the penultimate layer, giving us a 512-feature vector.

The experimental setup is identical to \S\ref{s:baseline_models}, except that for time considerations, we only consider CrossEntropyLoss.\footnote{Indeed, our UnbalancedCrossEntropyLoss code is not yet optimized, and slower than CrossEntropyLoss.} The test results for Emo8 classification are shown in \figref{fig:beyond_baseline_emo8_test}. Training results, as well as numerical training and test results, are included in \S\ref{a:additional_beyond_baseline}. A first observation is that improving upon the baseline appears non-trivial; except for VGG16, the obtained gains are modest. Second, the highest gains clearly come from adding facial emotion features. Third, even though adding EmoNet and OIToFER features separately has a positive effect for VGG16, adding both together does not result in a compounded improvement. Fourth, the added dimensionality of concatening features instead of predictions in the case of Places365 does not result in markedly different results, in some cases even leading to worse results. Finally, not a single stream combination resulted in improved performance for CLIP and DINOv2, with the best VGG16 results nearing CLIP/DINOv2 performance.

\section{Discussion}
\label{s:discussion}
\paragraph{Findings} The analysis of our dataset shows the annotations to conform to expectations, with Valence and Arousal values following the expected trends. Furthermore, when annotators disagree on the emotion label, they tend to choose nearby emotions in PWoE nonetheless. Our experiments show that, for the Emo8 prediction task on our dataset, modern ViT models do not seem to really outperform older CNN architectures, with VGG16 even (slightly) outperforming DINOv2 when both baselines are augmented with Facial Emotion features. For Arousal and Valence prediction however, the ViT models are clearly superior.

\paragraph{Limitations}
1) While images in our private set have multiple annotations, we have followed the approach of Emotic \cite{EMOTIC} and gathered only a single annotation per image in our public set. This choice has allowed us to gather a larger data set, but may cause concerns about reliability. These concerns are alleviated by the clear tendency observed on the private set toward similar emotions in case of multiple labels (\S\ref{a:inter_ann}), combined with trained models exhibiting this same tendency to strongly favor nearby emotion leaves when deviating from the annotation (\S\ref{a:disc_single_anns}). In short: the models trained using single annotations showed similar statistics to the human multi-label annotations. 2) Concerning potential biases in the images themselves, as they were scraped from the internet the dataset inherits the same biases the internet exhibits. In particular, we have not performed any analysis concerning potential representation issues. As such, there is an unverified possibility that models trained on our dataset wrongly associate ``negative'' emotions more strongly with certain minority groups. 3) Since legal issues (see \S\ref{a:legal_note}) prevent us from sharing the actual images, we had to resort to sharing URLs. While URLs can break, we mitigate this risk by offering multiple different URLs for the same image where possible.

\paragraph{Impact Statement}
This paper presents work whose goal is to advance the field of Machine Learning, Psychology and Psychiatry. Our own interest lies with non-commercial applications with respect to the understanding of Emotion Recognition and Social Cognition in individuals, and how these can be affected by neurological conditions. In particular, we hope that our (future) work will be of help in assisting people with impaired Social Cognition to navigate life.

Nevertheless, the data, and possible future Machine Learning advances inspired by it, could very well lead to commercial (e.g., personalized ads tailored to one's mood) and surveillance (e.g., general crowd monitoring, detection of aggression within crowds, etc.) applications that we strongly feel warrant a public debate with regard to their desirability, and even legality.

\paragraph{Conclusion}
We present FindingEmo, a dataset of 25k image annotations for Emotion Recognition that goes beyond the traditional focus on faces or single individuals, and is the first to target higher-order social cognition. The dataset creation process has been discussed in detail, and the annotations have been shown to align with expectations. Baseline results are presented for Emotion, Arousal and Valence prediction, as well as first steps to go beyond the baseline. These results show the dataset to be complex, and the tasks hard, with even modern models like CLIP and DINOv2 struggling. This suggests that in order to solve these tasks, novel Machine Learning roads might need to be explored. Our annotation interface and code for model training are made open source.

\begin{ack}
This work was funded by KU Leuven grant IDN/21/010.

We are grateful to dr. Simon Vandevelde for providing us with the dataset name.
\end{ack}

\bibliographystyle{ieeetran}
\bibliography{findingemo_v2}

\newpage
\appendix
\section{Appendix}

\subsection{Dataset Logo}
\label{a:dataset_logo}
\begin{figure}[ht]
\begin{center}
\centerline{\includegraphics[width=5cm]{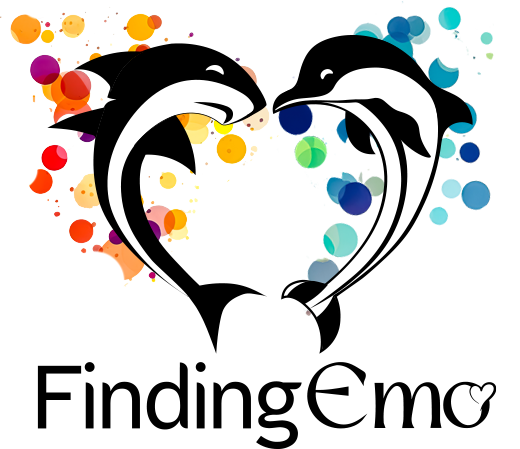}}
\caption{Logo for the FindingEmo dataset.}
\label{fig:logo}
\end{center}
\end{figure}

The logo of the dataset is depicted in \figref{fig:logo}.

\subsection{Legal Compliance}
\label{a:legal_note}
Concerning the legal status of the dataset, two question arise: 1) are we allowed to share URLs to (potentially) copyrighted content, and 2) are we allowed to use (potentially) copyrighted material to train our models?

With regard to 1, we verified this with copyright experts at our institute who assured us that this is legal. With regard to 2, we point to Title II, Article 3, ``Text and data mining for the purposes of scientific research'',  of the so-called InfoSoc Directive \footnote{\url{https://eur-lex.europa.eu/legal-content/EN/TXT/?uri=CELEX\%3A32019L0790}}, which provides an exception to copyright obligations for (members of) research organisations. As members of KU Leuven, we fall under this law. If you are not a member of a European research or cultural heritage institution, you will need to check with your local regulation whether or not you have the right to use this material for research purposes.

We are the rightful owners of the annotations, so no potential copyright issues arise for this data. We expressly distribute the dataset under a \emph{non-commercial} CC BY-NC-SA 4.0 license.

\subsection{Additional Annotation Dimensions}
\label{a:ann_add_dims}
These are the remaining annotation dimensions that were not mentioned in the main text for brevity.

\paragraph{Age group}
Users had to tick one or more boxes from ``Children'', ``Youth'', ``Young Adults'', ``Adults'' and ``Seniors'', indicating the age groups present in the image.

\paragraph{Deciding factor(s) for emotion}
Users had to tick one or more boxes from ``Neutral'', ``Body language'', ``Conflict context vs. person'', ``Facial expression'' and ``Staging'', indicating what prompted them to choose for a particular emotion.

\paragraph{Ambiguity}
Lastly, users could indicate by means of an integer scale $[0, 1, \dots, 6]$ how ambiguous the emotional content exhibited by the entire photograph was, or alternatively, how much difficulty they had in annotating the picture.

\subsection{Details of the Dataset Creation Process}
\label{a:dataset_creation}
This section describes in more detail the two phases in the dataset creation process introduced in \S\ref{ss:dataset_creation_process}.

\subsubsection{Phase 1: Gathering Images}
\label{a:sss:gathering_imgs}
Phase 1 consisted in building a customized, Python-based DuckDuckGo\footnote{\url{https://www.duckduckgo.com}} image scraper, programmed to generate random image search queries as follows. Three sets of keywords were defined: one containing a diverse set of emotions; one containing social settings and environments (e.g., `birthday', `workplace', etc.); and one referring to humans (e.g., `people', `adults', `youngsters', etc.).\footnote{The full list of keywords is available from our code repository.} By taking all possible combinations of the elements in these sets, the system generated a multitude of queries, such as, e.g., ``happy youngsters birthday''. The first $N$ results were then retrieved and filtered to exclude a number of manually blacklisted domains (e.g., stock photography providers) and by image size. Query results that passed the filtering steps were downloaded.

We started with $N=500$ and image width $800$px $< w < 1600$px, and later extended this to $N=1000$ and $800$px $< w < 3200$. Obviously, not all downloaded images satisfied our criterion of depicting multiple people in a natural setting. Hence, as a further filtering step, one of the authors annotated 3097 images as either ``keep'' (useful) or ``reject'' (no use). These images were used in a random 80/20 split to train a CNN to perform the same task, achieving an accuracy of 77.6\%. This model was used to further filter downloaded images, in particular to identify spurious images such as, e.g., drawings, images with lots of text, etc.: if the CNN labeled the downloaded image as ``reject'', the image was discarded. If the downloaded image was labeled as ``keep'', it entered the pool of images that could be selected for annotation.

In total 1,041,105 images were collected.

\subsubsection{Phase 2: Gathering Annotations}
\label{a:sss:gathering_anns}
The annotations were gathered using a custom web interface written in Python, HTML and JavaScript. Annotators were recruited through the Prolific platform. For this, a job would be created, which we refer to as a ``run'', to which users could subscribe. After doing so, they received a URL that allowed them to log on to our system and, after agreeing to an Informed Consent clause, perform the annotations. First, users were presented with detailed instructions, a copy of which are provided in \S\ref{a:ann_instructions}, after which the data collection proper began. To be able to monitor the process closely, and to cope with hardware limitations of our server, we opted to only perform runs with a limited number of participants, most often 10 or 15. For each run, the Prolific user selection criteria were the same: fluent English speaker, (self-reported) neurotypical, and a 50/50 split male/female.

In total, annotations were collected over 51 runs. Candidates were informed of an expected task duration of 1h, including reading the instructions, and offered a $\pounds$10 reward. Analysis of the durations (see \S\ref{a:duration_analysis}) show our time estimation to be fair. We spent a total of $\pounds$10k, which includes annotators whose contributions were filtered out, and most importantly, Prolific fees and taxes.

A screenshot of the interface is included in \S\ref{a:ann_interface}. The interface presents users with images on the left side, and dimensions to annotate on the right side. At the top left, users are presented with two buttons: one to skip an image if they so wish, and one to save the current annotation and move on to the next image.

Upon being presented an image, the first choice users needed to make was, just like the filtering CNN, whether to ``keep'' or ``reject'' the image, according to the provided instructions. Essentially, users were asked to reject images that contained no people, were watermarked, were of bad quality, etc. If users opted to ``reject'' an image, no further annotation was needed. This step was needed to further filter images that passed through the CNN. If the choice was ``reject'', no further action (besides saving) was required. Optionally, users could choose to select one of several tags indicating why they opted to reject the image from ``Bad quality photo'', ``Copyright'', ``Watermark'', ``No interaction'', ``No people'', ``Text'' and ``Not applicable''. Each user was asked to annotate 50 ``keep'' images; ``rejects'' did not count towards the total goal. Despite this, some users still performed full annotations on images they rejected. If users opted to ``keep'' the image, they were expected to annotate all other dimensions as well.

Although the frontend (i.e., user interface) remained essentially unchanged, the backend underwent some changes as annotations were collected, and some lessons were learned, which we discuss here.

\paragraph{Initial iteration}
Initially, an image was randomly selected from the corpus, and processed by an updated ``keep/reject'' CNN (see \S\ref{a:sss:gathering_imgs}) with an accuracy of 83.6\%. If the ``keep'' probability $p_k$ was $< 0.75$, a new random image would be selected and tested, until one was found with $p_k \geq 0.75$. If this image had already been annotated, the process would start over, until a valid image was found, which would then be shown to the annotator.

\paragraph{Second iteration}
At first, the annotating of all dimensions was not enforced; users could select the ``keep'' checkbox, save the annotation without annotating anything else, and move on to the next image. Most did their job diligently, but nevertheless we opted to update the interface to require all dimensions be annotated in case of a ``keep'', before the ``Save'' option became available. This frequently prompted messages from users complaining the ``Save'' option was not available to them. A further update explained this to users who prematurely clicked on the ``Save'' button.

\paragraph{Third iteration}
Over the course of the first few thousand annotations, it became clear that two emotion leaves were particularly overrepresented, namely ``joy'' and ``anticipation'', respectively accounting for $35.9$\% and $23.0$\% of all annotations by the time of Run 9. In an attempt to counter this, we came up with the following system.

Besides the ``keep/reject'' CNN, we trained a second CNN to predict the Emo8 label. We then first computed all ``keep/reject'' predictions for all images in the corpus, and followed this up by predicting Emo8 labels for all ``keep''-labeled images. Upon starting the annotation server, these predictions are loaded into memory. When selecting an image to show to a user, first an emotion label is chosen, with odds inversely proportional to the number of images that were tagged (by the CNN) with a certain label. Second, out of all images tagged with this label, one that had not previously been annotated by an annotator would be chosen. The CNN used to make the predictions was retrained at several steps along the annotation gathering process. Using this system, we managed to decrease ``joy'' down to $28.4$\%, and up ``sadness'' from $6.3$\% to $10.5$\%.

\subsection{Annotation Interface}
\label{a:ann_interface}
A screenshot of the annotation interface is shown in \figref{fig:ann_interface}.

\afterpage{\clearpage
\begin{landscape}
\vspace*{\fill}
\begin{figure}[!htbp]
\begin{center}
\centerline{\frame{\includegraphics[width=21cm]{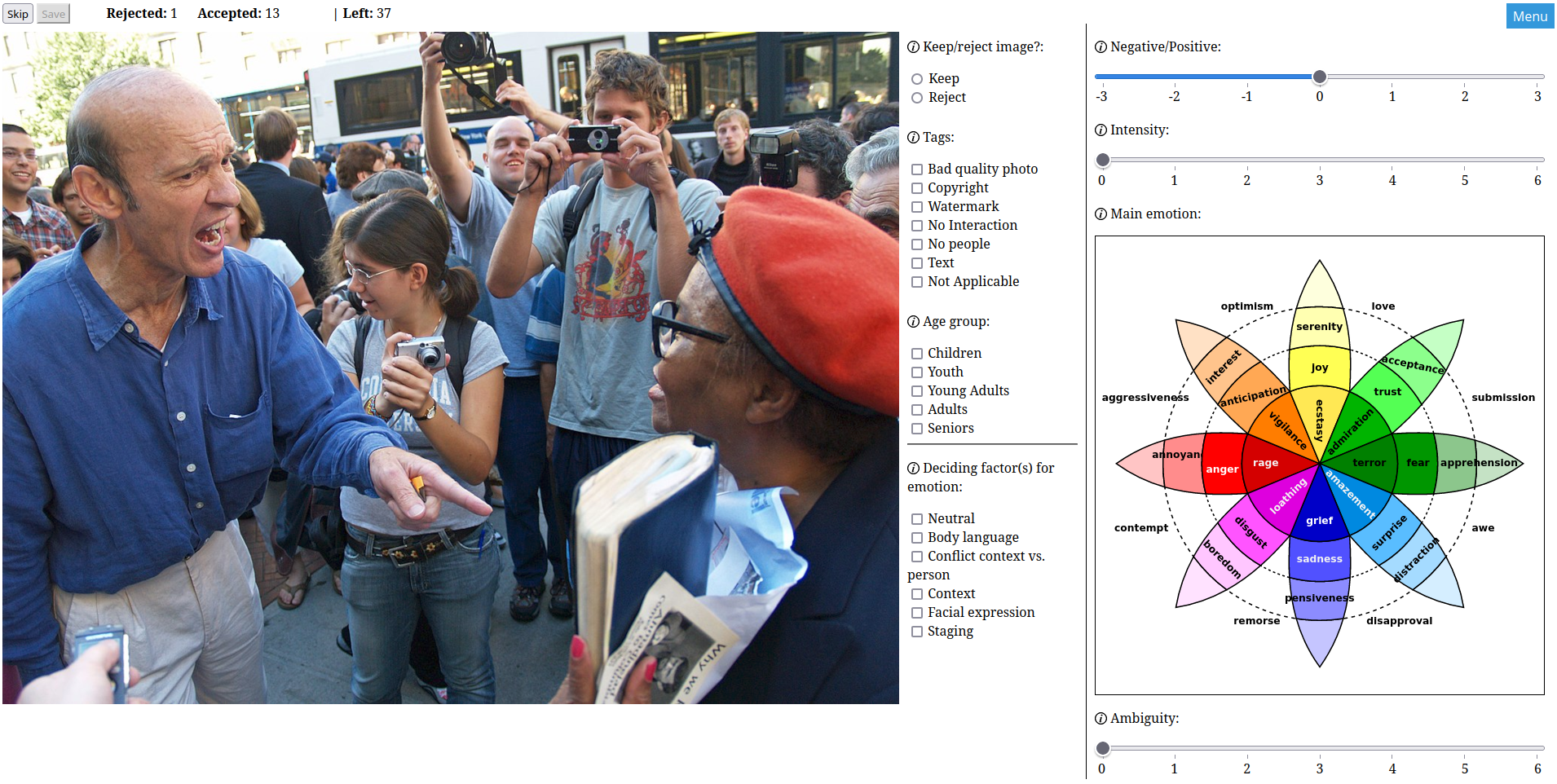}}}
\caption{A screenshot of the annotation interface. Displayed photo by David Shankbone, source: \href{https://commons.wikimedia.org/wiki/File:Anger_during_a_protest_by_David_Shankbone.jpg}{WikiMedia}.}
\label{fig:ann_interface}
\end{center}
\end{figure}
\vfill
\end{landscape}
}

\subsection{Copy of the Annotator Instructions}
\label{a:ann_instructions}
Welcome

It is recommended to set your browser to ``full-screen'' mode. Typically, this mode can be toggled by using the `F11' key.

This interface was designed for screen resolutions with a width of 1920 pixels. In case your screen has a higher/lower resolution, the interface should automatically resize itself so as to fully fit on your screen, but this might come at the price of reduced image sharpness.

Thank you for your willingness to participate in this annotation task!

In this experiment, you will be expected to annotate 50 ``good'' images, i.e., annotated as ``Keep'', after which you will receive a URL that will direct you to the Prolific completion page for this task. Please take the time to read these annotation instructions before continuing.

Note that if for any reason you get logged out at some point, you should be able to log back in using the same URL provided to you by Prolific, and pick up right where you left.

We want to build a database of photographs with an emotional content. You will be shown randomly selected images from a large corpus, and we ask you to evaluate photographs regarding 2 consecutive issues.

First, regardless of the emotional content, all photographs should adhere to the following criteria:
\begin{itemize}
\item Each photograph must display a realistic situation, e.g., no drawings, no watermark, no fantasy content (i.e., digitally manipulated photos), no horror, etc.
\item The formal quality of the photograph should be sufficient, i.e., no fuzzy/blurry photographs.
\item Each picture must display at least 2 people that are clearly visible. Alternatively, if only one person is shown, but this person is clearly a part of a larger context, the image can also be suitable.
\item The main feature of the photograph must not consist of a textual element. For instance, if a cardboard displaying ‘stop racism’ is a central feature of the picture, the picture is not suitable.
\end{itemize}

If an image does not adhere to each of these criteria, or you are not certain, please rate it as not suitable by choosing the ``Reject'' option. Else, mark it as ``Keep'', in which case all other dimensions, except for ``tags'', need to be annotated before you can proceed! Even if you want to keep the default value of a slider, you still need to click the slider first.

Images can further be described by a number of tags:
\begin{itemize}
\item Bad quality photo: when a picture is too blocky/blurry.
\item Copyright: a copyright, contrary to a watermark, is not repeated but appears only once. Typically, this leads to the picture being rejected, unless possibly the copyright is only small in size and could be cropped out without losing the essence of the picture.
\item Watermark: a watermark is a specific pattern, typically containing the name of the copyright holder, that is repeated over an entire image.
\item No interaction: the people in the picture don't have a direct interaction.
\item No people: the picture does not depict any people.
\item Text: the image contains a lot of text, either typeset on top of it, or present on, e.g., banners held by subjects depicted in the picture. If the text is typeset, this is disqualifying (i.e., the picture is rejected). If the text is present in the picture itself, it is disqualifying if it is too prominent. Use your own discretion to determine what is ``too prominent'' and what is not. A good rule of thumb is: if your attention is immediately drawn to the textual elements when viewing the picture, then it is too prominent and the picture is disqualified.
\item Not Applicable: typically used for images that are actually a collage of more than one photo, or that are rejected but don't fit any of the other tags.
\end{itemize}

If a photograph is not rated as suitable (i.e., ``Reject''), no further assessment is required; click ``Save'' to proceed to the next paragraph. Else, for ``Keep'' or ``Uncertain'' photos, you are also expected to annotate the age group of the main participants in the picture. These labels are of course not clear cut; feel free to use your own discretion as to which label applies best.

Second, we want you to focus on the emotional labelling of the photographs. Concretely, we ask you to annotate the image on a number of dimensions

We ask you to indicate the emotional characteristic of the ENTIRE SCENE displayed in the photograph, independent of your own political/religious/sexual orientation. So a black lives matter protest is typically negative (= the participants are not happy) independent of whether you support BLM. Specifically, we ask you to rate the valence (``Negative/Positive'') of the overall emotional gist of the photograph on a 7-point Likert scale from negative (-3) over neutral (0) to positive (+3), and also the intensity, ranging from not intense at all (0) to very intense (6) by using the appropriate sliders.

We also ask to indicate an emotional label by means of a mouse click on an emotion wheel called ``Plutchik's Wheel of Emotions''. If you can’t find the perfect emotional label then you choose the `next best thing', i.e., the one that reflects it most. In case no particular emotion fits, i.e., the participants all display a neutral expression, you can opt to select no emotion, although such cases are expected to be rare. For a more detailed description of each emotion depicted in this wheel, see, e.g.,  \url{https://www.6seconds.org/2020/08/11/plutchik-wheel-emotions/}. Additional info for each emotion will be displayed when hovering over its corresponding cell.

Please also rate how straightforward the emotional content that is exhibited by the entire photograph is using the scale indicated with ``Ambiguity''. For instance, if there are approximately as much emotionally positive as emotionally negative cues in the photograph, the emotional content would not be clear (6), while only positive cues or only negative cues would result in a very high clarity (0).

Finally, the options under the ``Deciding factor(s) for emotion'' header ask which aspects of the photo influenced you most when assessing the emotion, i.e., facial expressions, bodily expressions, the type of interaction (`Staging') among the persons (e.g., fighting, dancing, talking), type of context (e.g., wedding, funeral, protest, etc.), objects in the photograph (e.g., gun, chocolate) or a possible conflict between context and person(s) (i.e., somebody exuberantly laughing at a funeral). If none of these apply, and/or the emotion is rather neutral, the ``Neutral'' tag can be used, although just as for the emotion case, we expect these occasions to be rare.

If for some reason you would rather not annotate the current image being served to you, you can press the ``Skip'' button to be served a new picture and have the annotation interface be reset, without your current settings being saved.

If on the other hand you are happy with your current annotation, press ``Save'' to let it be saved and move on to the next image. If this button is greyed out, this means you have not yet annotated all necessary dimensions. Once you have reached the required number of annotations, you will automatically get to see the URL that will direct you to the Prolific completion page for this task.

At the top of this screen, you can see your annotation statistics: ``Rejected/Accepted'' = how many images you marked ``Reject'' and ``Keep'' respectively, and ``Left'' = number of ``Keep'' images left to annotate.

You can always check these instructions again whilst annotating by clicking the -icon next to each criterium. (Click once more to close the infobox again.)

\subsection{Task Duration Analysis}
\label{a:duration_analysis}
\begin{figure}[!htb]
\begin{center}
\centerline{\includegraphics[width=8cm]{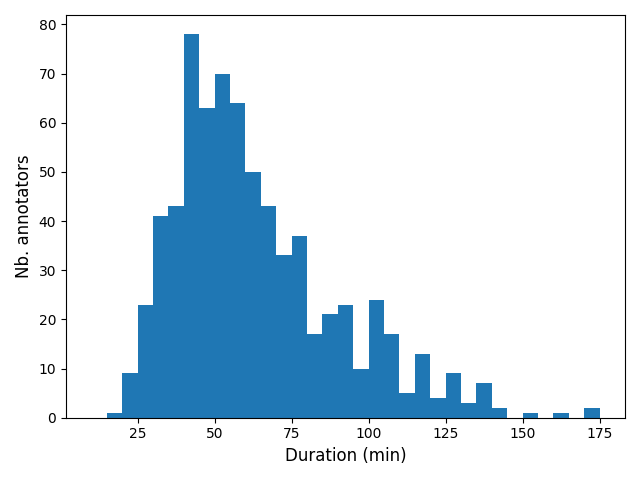}}
\caption{Distribution of minutes taken to complete the task. The plot does not include 7 outliers.}
\label{fig:ann_duration}
\end{center}
\end{figure}

A histogram of time taken per annotator to complete the task is shown in \figref{fig:ann_duration}. These are the durations as reported by Prolific. An important remark to make is that for Prolific users, the clock starts ticking once they subscribe to a job. By default, per the Prolific rules, for a job expected to take 1h users are allowed a maximum of 140 minutes to complete the job. It appears that many users subscribe to a job, and then leave their browser tab open for a while before starting the job proper. (Some never start, leading to a time-out.) Taking this into account, the shown distribution is a ``pessimistic'' picture, including many idled minutes. The average time taken per user, including users that were ultimately filtered out of the dataset, was 64 $\pm$ 27 minutes. With all of the above in mind, we conclude our alloted time was fair.

A small negative correlation manifests between the task completion time and the annotator score (SpearmanR$=-0.122$, $p=0.002$ for $s$, SpearmanR$=-0.086$, $p=0.029$ for $s_{alt}$).

\subsection{Annotator Statistics}
\label{a:ann_stats}
Annotations were collected from 655 annotators. Prolific provided us with anonymized personal data, except for 1 user. Not all datapoints are available for all users.

\begin{figure}[!htb]
\begin{center}
\centerline{\includegraphics[width=17cm]{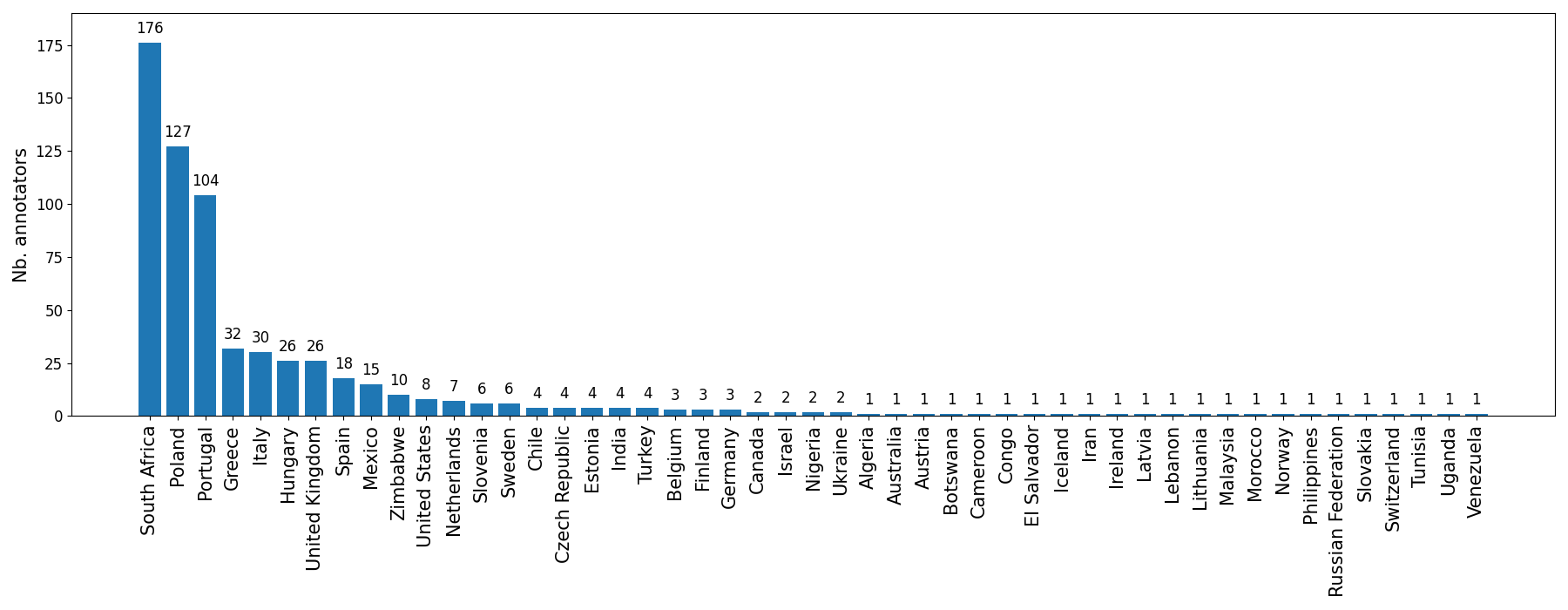}}
\caption{Distribution of country of origin of 651 annotators.}
\label{fig:distri_ann_country}
\end{center}
\end{figure}

\begin{figure}[!htb]
\begin{center}
\centerline{\includegraphics[width=17cm]{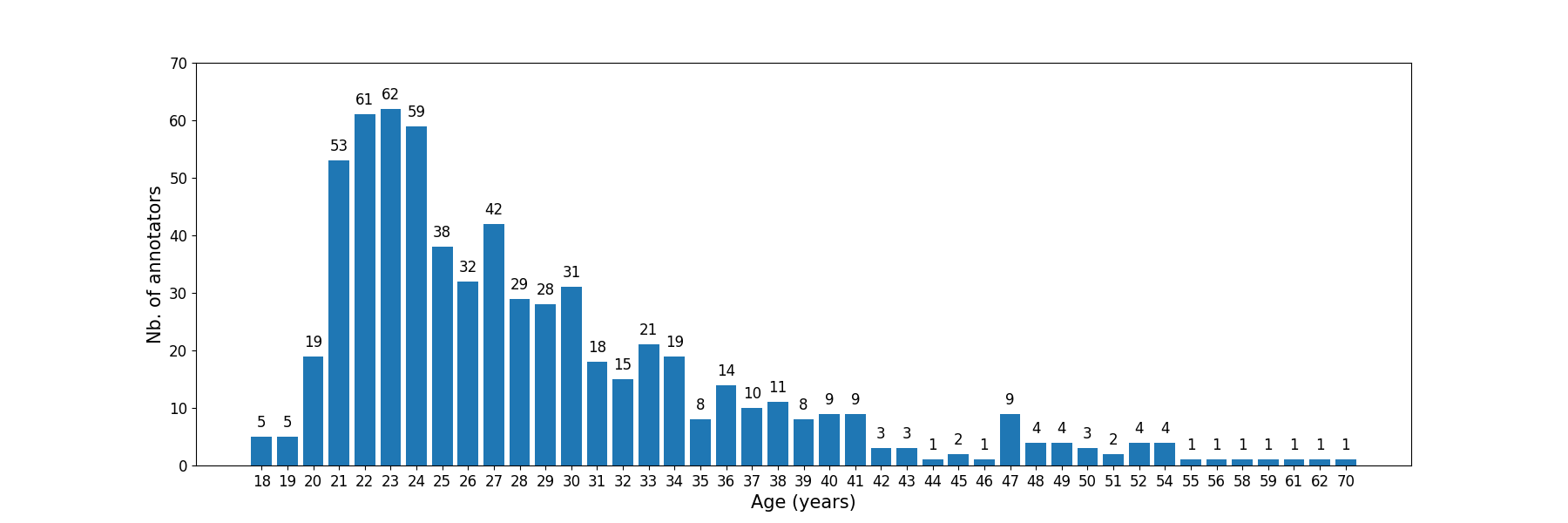}}
\caption{Distribution of age of 653 annotators.}
\label{fig:distri_ann_age}
\end{center}
\end{figure}

Of the annotators, 337 are male, 317 are female, and 1 unknown. 651 annotators were spread over 49 countries, with country for the remaining 4 unknown. Most popular were South Africa (176 annotators), Poland (127 annotators) and Portugal (104 annotators). From there, numbers drop rapidly, with follow-up Greece accounting for only 32 annotators. The full distribution of annotators per country is shown in \figref{fig:distri_ann_country}. The age distribution of the 653 users who shared that info is shown in \figref{fig:distri_ann_age}, indicating a large bias towards the early 20's.

\subsection{Inter-annotator Agreement}
\label{a:inter_ann}
Recall from \S\ref{s:dataset_desc} that we hold a private set of 1,525 images that have each been annotated by multiple users\footnote{This set does not include the fixed overlap images.}, amounting to a combined 6115 annotations. \tabref{tb:nb_imgs_per_nb_anns} shows how many images have been annotated by $N$ different annotators. Of these, 1294 images have a majority of ``Keep'' annotations, 137 are mainly ``Reject'' and 94 are undecided.

We do not report the often used Cohen's Kappa and/or Krippendorf's Alpha scores, as these metrics are only meaningful when most pairs of annotators have both annotated a substantial set of shared images. In our case, however, by design, the number of images that have been annotated by any two annotators is low (1 or 2 at most, and very often zero). As such, we feel these metrics are not applicable. We explicitly opted to have a large number of annotators annotate a small number of images each, in order to have the annotations better be a reflection of ``the population at large'', rather than of a few annotators.

\begin{table}[!htb]
	\centering
	\caption{Number of images (``\# imgs.'') that have been annotated by $N$ different annotators (``\# ants.'').}
	\label{tb:nb_imgs_per_nb_anns}
	\begin{tabular}{@{}l|lllllll@{}}
\# ants. & 2 & 3 & 4 & 5 & 6 & 7 & 8 \\
\# imgs. & 245 & 328 & 283 & 524 & 127 & 17 & 1
	\end{tabular}
\end{table}

\begin{figure}[!htb]
\begin{center}
\centerline{\includegraphics[width=17cm]{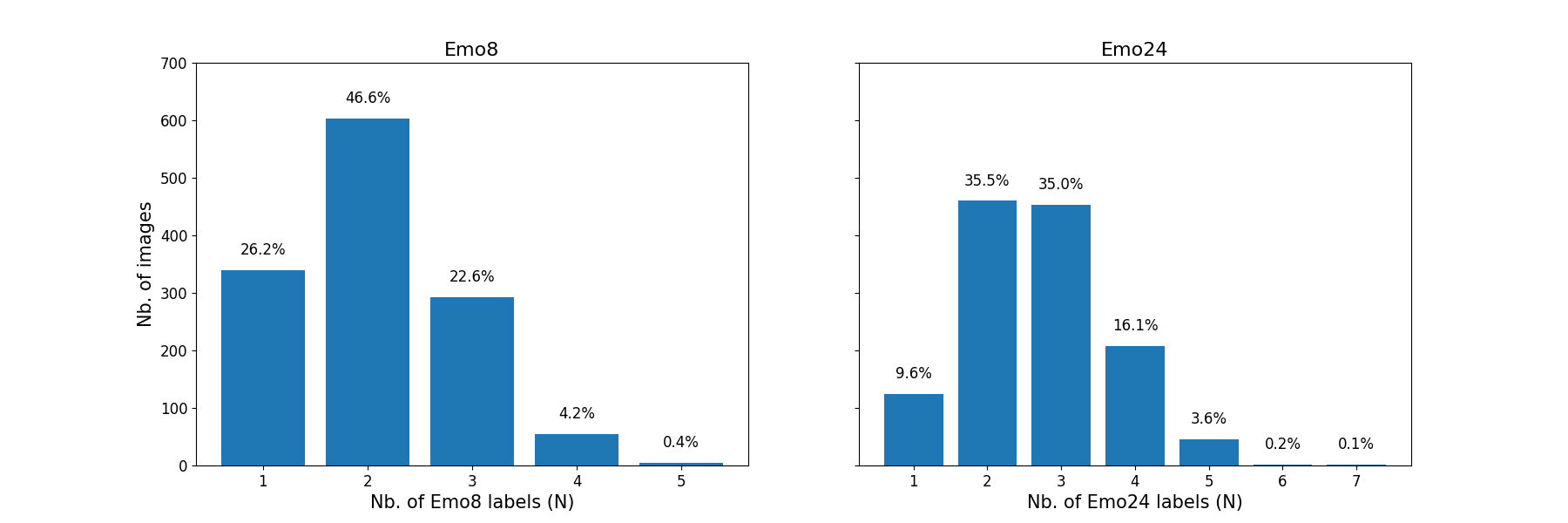}}
\caption{Number of images with $N$ different Emo8 and Emo24 labels. The y-axis is shared between both plots.}
\label{fig:hist_multiann_nblabels}
\end{center}
\end{figure}

Focusing on the 1294 ``Keep'' images, \figref{fig:hist_multiann_nblabels} shows how many images have been annotated with $N$ different emotion labels, both for Emo8 and Emo24 labels. For 26.2\% of images, all annotators chose the same emotion leaf, and 46.6\% were annotated with 2 different Emo8 labels. For the finegrained Emo24 labels, 80.1\% of images have been annotated with a maximum of 3 different labels.

\begin{figure}[!htb]
\begin{center}
\centerline{\includegraphics[width=17cm]{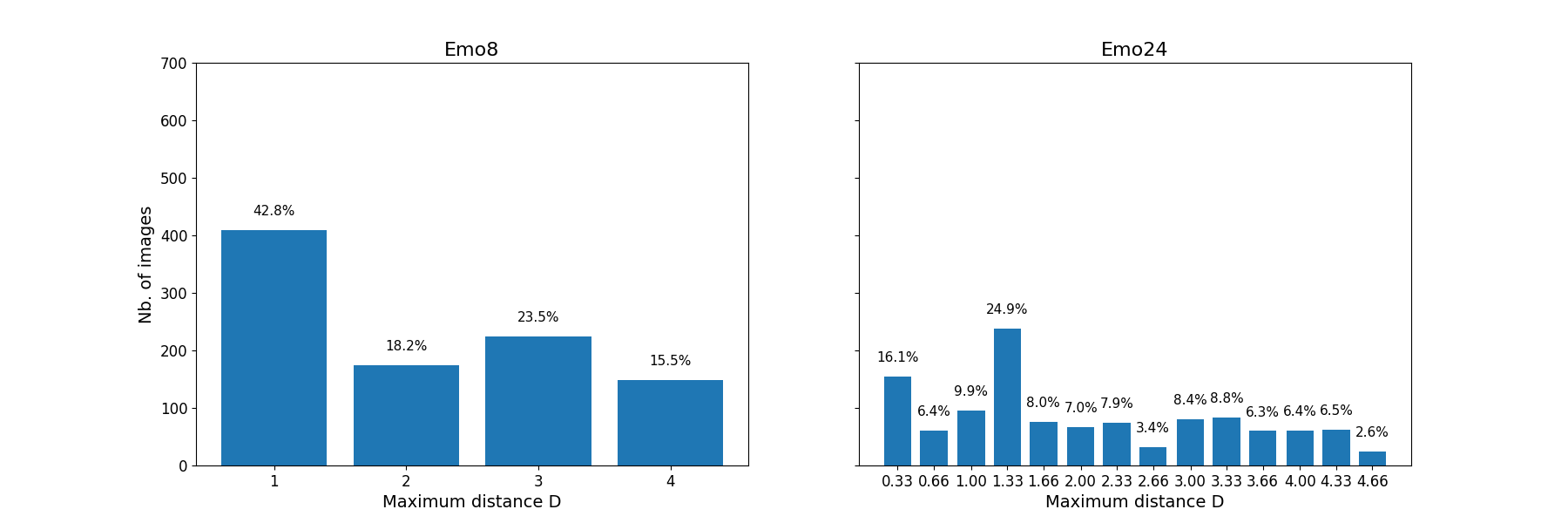}}
\caption{Number of images with a maximum distance $D$ between their Emo8 and Emo24 labels, for images annotated with more than one label. The y-axis is shared between both plots.}
\label{fig:hist_multiann_maxdist}
\end{center}
\end{figure}

\begin{figure}[!htb]
\begin{center}
\centerline{\includegraphics[width=7cm]{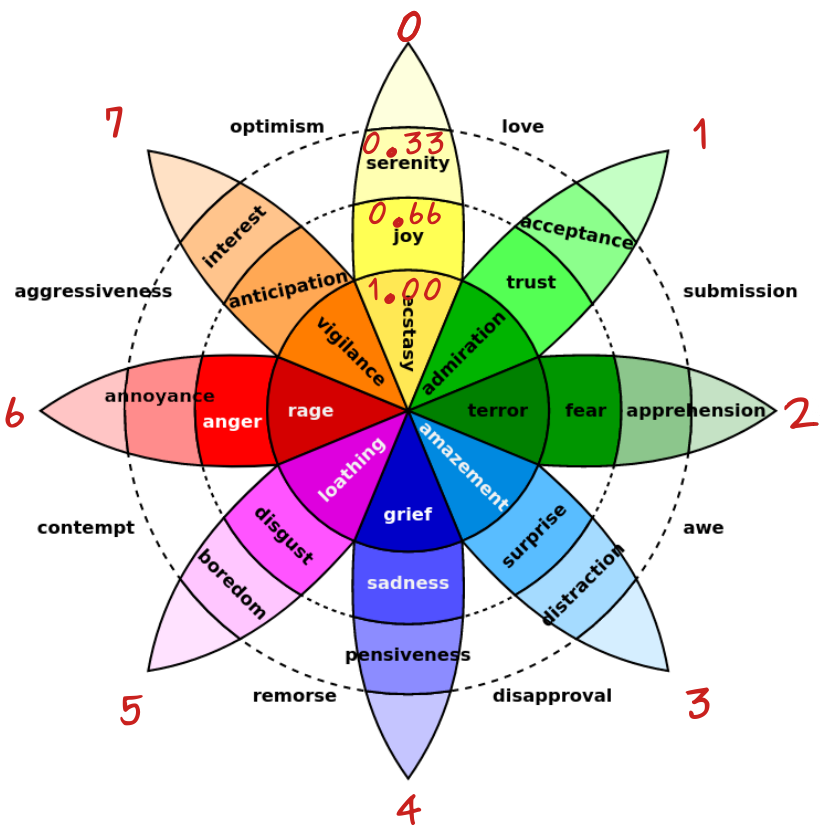}}
\caption{Plutchik's Wheel of Emotions: ordinals of emotions. The outer numbers represent the ordinal of the leaf, the numbers within the upper central leaf the ordinals of the emotions within a leaf. E.g., ``joy'' = 0.66 and ``boredom'' = 5.33. The distance between them then becomes 3.33, being the sum of the distance between the leaves (3) and the ``intra-leaf'' distance (0.33).}
\label{fig:pwoe_dist}
\end{center}
\end{figure}

Turning to the question of how different the separate emotion labels for a same image are, \figref{fig:hist_multiann_maxdist} shows the distribution of maximum distance between labels, for images annotated with more than one label. The distances for 24 emotions are computed by also giving an ordinal to each emotion within a leaf, as shown in \figref{fig:pwoe_dist}. No less than 42.8\% of the times an image has been annotated with more than one Emo8 label, those labels represent adjacent emotion leaves, while in 15.5\% of the cases they represented opposite leaves, most often the paris (``anger'', ``fear'') and (``anticipation'', ``surprise'').

\begin{figure}[!htb]
\begin{center}
\centerline{\includegraphics[width=8cm]{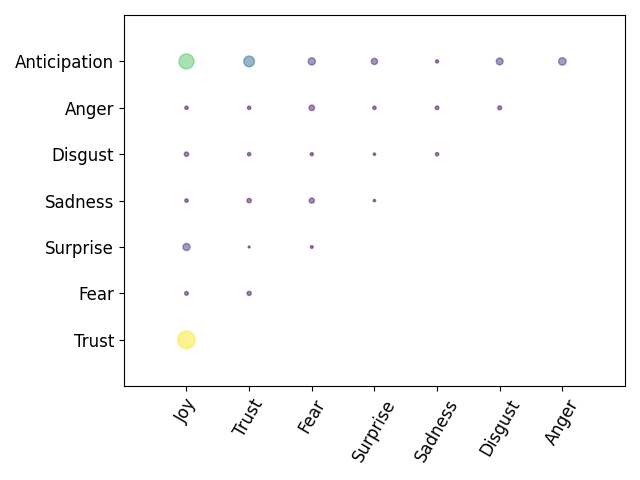}}
\caption{Prevalence of Emo8 label pairs for images annotated with 2 labels. The bigger the disc, the more often the pair appears in the dataset.}
\label{fig:corr_multiann_emopairs}
\end{center}
\end{figure}

To get a better idea of what Emo8 labels often appear together, we focused on images with 2 Emo8 labels, and plotted how often each emotion pair occurs. The result is shown in \figref{fig:corr_multiann_emopairs}, demonstrating the pairs (``Joy'', ``Anticipation''), (``Joy'', ``Trust) and (``Aniticipation'', ``Trust'') make up the bulk of the pairs. As for opposite emotions, the pairs (``Anticipation'', ``Surprise'') and (``Anger'', ``Fear'') appear markedly more ofthen than (``Joy'', ``Sadness'') and (``Disgust'', ``Trust'').

\begin{figure}[!htb]
\begin{center}
\centerline{\includegraphics[width=17cm]{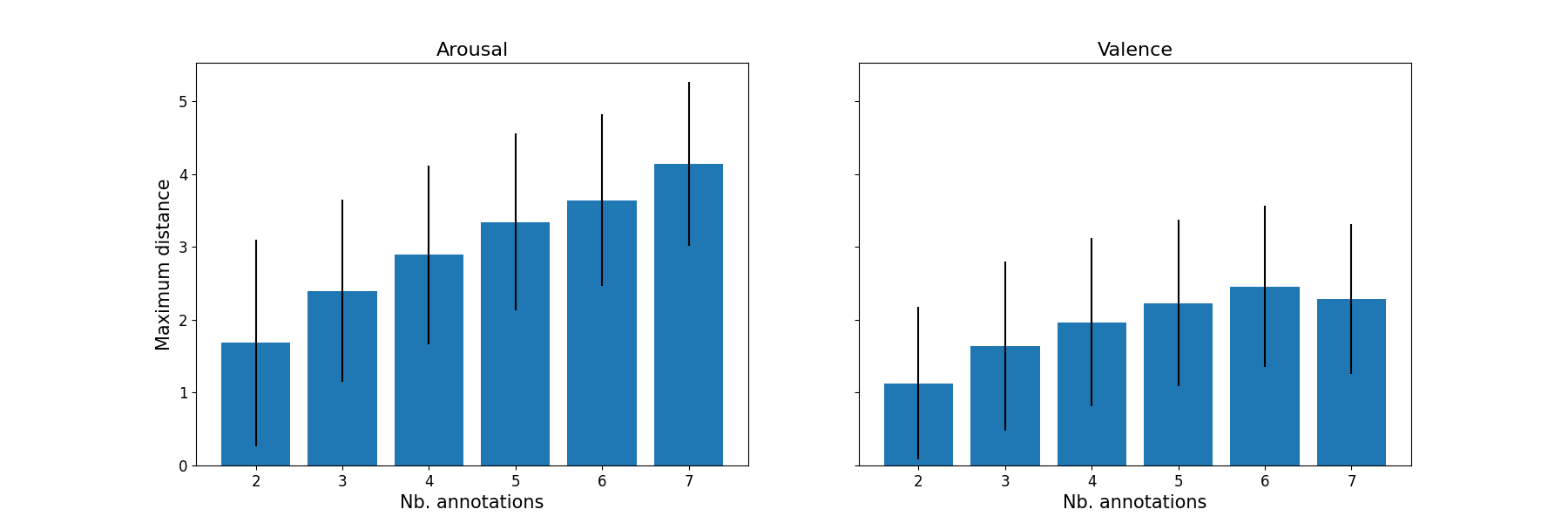}}
\caption{Distribution of maximum distance between Arousal and Valence annotations as a function of the number of annotations per image. The y-axis is shared between both plots.}
\label{fig:hist_multiann_maxdist_aro_val}
\end{center}
\end{figure}

To analyze the Arousal and Valence values, we compute the maximum distance between annotated values for both dimensions over all ``keep'' images. For Arousal, the average maximum distance is 2.7 $\pm$ 1.4, while for Valence this is 1.8 $\pm$ 1.2. This suggests that people agree much more on the Valence dimension, than they do on the Arousal dimension. This is confirmed when we compute the average maximum distance values as a function of the number of annotations for a given image, the result of which is shown in \figref{fig:hist_multiann_maxdist_aro_val}. For Arousal, a clear increasing maximum distance trend is visible with a stable standard deviation, going from $\pm$1.75 to more than 4. For Valence annotations on the other hand, the maximum distance appears to plateau at close to 2.

\subsection{Extra Dataset Analysis}
\label{a:extra_data_analysis}
\begin{figure}[h]
\begin{center}
\centerline{\includegraphics[width=17cm]{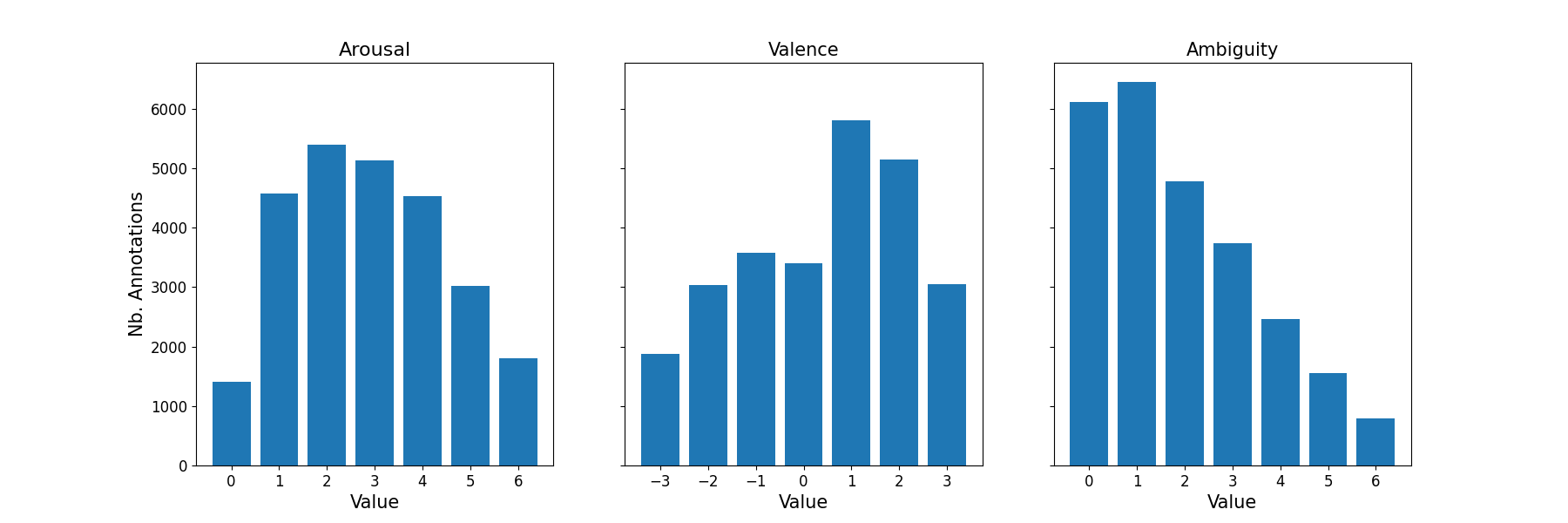}}
\caption{Distribution of Arousal, Valence and Ambiguity annotations. The y-axis is shared between all plots.}
\label{fig:hist_aro_val_amb}
\end{center}
\end{figure}

Histograms showing the distribution of Arousal, Valence and Ambiguity annotation values for the public dataset are depicted in \figref{fig:hist_aro_val_amb}. Annotation statistics per Emo24 emotion are collected in \tabref{tb:emo24_ann_stats}. The table is made up of three rows, each row corresponding to a ring in Plutchik's Wheel of Emotions, from the top row corresponding to the outer (least intense) ring, to the bottom row corresponding to the inner (most intense) ring. The annotations follow this ordering, with average Arousal annotations consistently increasing from least to most intense emotion ring. Valence annotations follow suit, either increasing for positive emotions, or decreasing for negative emotions. The sole exception to this rule is center ring ``Disgust'' having a slightly lower average Valence rating (-1.62) than the inner ring ``Loathing'' (-1.57).

\afterpage{
\begin{landscape}
\vspace*{\fill}
\begin{table}[!htb]
	\centering
	\caption{Average Arousal, Valence and Ambiguity annotation values for the public set, per emotion. Emotions are grouped per ``ring'' in Plutchik's Wheel of Emotions: from outer, least intense ring (top row) to inner, most intense ring (bottom row). The percentage of total annotations per emotion is shown in square brackets. Format: $x.xx^{y.yy}$ should be read as average = $x.xx$, standard deviation = $y.yy$.}
	\label{tb:emo24_ann_stats}
	\begin{tabular}{@{}l|llllllll@{}}
		 & Serenity & Acceptance & Apprehension & Distraction & Pensiveness & Boredom & Annoyance & Interest \\
		Nb. & \phantom{$-$}$1972$ [7.6\%] & \phantom{$-$}$1388$ [5.4\%] & \phantom{$-$}$1400$ [5.4\%] & \phantom{$-$}$ 356$ [1.4\%] & \phantom{$-$}$ 706$ [2.7\%] & \phantom{$-$}$ 587$ [2.3\%] & \phantom{$-$}$1008$ [3.9\%] & \phantom{$-$}$2688$ [10.4\%] \\
		Arousal & \phantom{$-$}$2.24^{1.00}$ & \phantom{$-$}$2.21^{1.05}$ & \phantom{$-$}$2.73^{1.12}$ & \phantom{$-$}$2.04^{1.11}$ & \phantom{$-$}$2.53^{1.17}$ & \phantom{$-$}$1.65^{0.94}$ & \phantom{$-$}$2.78^{1.02}$ & \phantom{$-$}$2.03^{0.99}$ \\
		Valence & \phantom{$-$}$1.46^{1.00}$ & \phantom{$-$}$1.14^{1.05}$ & $-0.96^{1.12}$ & $-0.06^{1.11}$ & $-0.90^{1.17}$ & $-0.36^{0.94}$ & $-1.20^{1.02}$ & \phantom{$-$}$0.79^{0.99}$ \\
		Ambiguity & \phantom{$-$}$1.83^{1.65}$ & \phantom{$-$}$2.00^{1.65}$ & \phantom{$-$}$2.26^{1.61}$ & \phantom{$-$}$2.71^{1.65}$ & \phantom{$-$}$2.33^{1.63}$ & \phantom{$-$}$2.30^{1.63}$ & \phantom{$-$}$2.21^{1.55}$ & \phantom{$-$}$2.16^{1.62}$ \\
		\midrule
		 & Joy & Trust & Fear & Surprise & Sadness & Disgust & Anger & Anticipation \\
		Nb. & \phantom{$-$}$3971$ [15.4\%] & \phantom{$-$}$1145$ [4.4\%] & \phantom{$-$}$ 679$ [2.6\%] & \phantom{$-$}$ 311$ [1.2\%] & \phantom{$-$}$1092$ [4.2\%] & \phantom{$-$}$ 248$ [1.0\%] & \phantom{$-$}$ 985$ [3.8\%] & \phantom{$-$}$1780$ [6.9\%] \\
		Arousal & \phantom{$-$}$3.04^{0.86}$ & \phantom{$-$}$2.57^{1.05}$ & \phantom{$-$}$3.76^{1.13}$ & \phantom{$-$}$2.76^{1.41}$ & \phantom{$-$}$3.44^{1.18}$ & \phantom{$-$}$3.31^{1.08}$ & \phantom{$-$}$3.99^{1.11}$ & \phantom{$-$}$2.58^{1.26}$ \\
		Valence & \phantom{$-$}$2.02^{0.86}$ & \phantom{$-$}$1.49^{1.05}$ & $-1.79^{1.13}$ & \phantom{$-$}$0.35^{1.41}$ & $-1.65^{1.18}$ & $-1.62^{1.08}$ & $-1.80^{1.11}$ & \phantom{$-$}$0.49^{1.26}$ \\
		Ambiguity & \phantom{$-$}$1.47^{1.60}$ & \phantom{$-$}$1.75^{1.59}$ & \phantom{$-$}$1.92^{1.53}$ & \phantom{$-$}$2.21^{1.70}$ & \phantom{$-$}$1.82^{1.61}$ & \phantom{$-$}$2.12^{1.59}$ & \phantom{$-$}$1.86^{1.63}$ & \phantom{$-$}$2.24^{1.55}$ \\
		\midrule
		 & Ecstasy & Admiration & Terror & Amazement & Grief & Loathing & Rage & Vigilance \\
		Nb. & \phantom{$-$}$1083$ [4.2\%] & \phantom{$-$}$1016$ [3.9\%] & \phantom{$-$}$ 322$ [1.2\%] & \phantom{$-$}$ 221$ [0.9\%] & \phantom{$-$}$ 867$ [3.4\%] & \phantom{$-$}$ 165$ [0.6\%] & \phantom{$-$}$ 446$ [1.7\%] & \phantom{$-$}$1433$ [5.5\%] \\
		Arousal & \phantom{$-$}$3.97^{0.93}$ & \phantom{$-$}$3.06^{1.09}$ & \phantom{$-$}$4.34^{1.37}$ & \phantom{$-$}$3.14^{1.31}$ & \phantom{$-$}$4.12^{1.29}$ & \phantom{$-$}$3.93^{1.41}$ & \phantom{$-$}$4.54^{1.38}$ & \phantom{$-$}$3.13^{1.41}$ \\
		Valence & \phantom{$-$}$2.27^{0.93}$ & \phantom{$-$}$1.70^{1.09}$ & $-2.03^{1.37}$ & \phantom{$-$}$1.51^{1.31}$ & $-2.01^{1.29}$ & $-1.57^{1.41}$ & $-1.92^{1.38}$ & \phantom{$-$}$0.22^{1.41}$ \\
		Ambiguity & \phantom{$-$}$1.52^{1.83}$ & \phantom{$-$}$1.84^{1.66}$ & \phantom{$-$}$1.69^{1.65}$ & \phantom{$-$}$2.13^{1.61}$ & \phantom{$-$}$1.46^{1.64}$ & \phantom{$-$}$2.07^{1.79}$ & \phantom{$-$}$1.80^{1.72}$ & \phantom{$-$}$2.02^{1.66}$ \\
	\end{tabular}
\end{table}
\vfill
\end{landscape}
}

\subsection{UnbalancedCrossEntropyLoss and WeightedMSELoss}
\label{a:uce_wmse}
\tabref{tb:baseline_cs_vs_uce} compares baseline results obtained using CrossEntropyLoss vs. UnbalancedCrossEntropyLoss for Emo8 classification, and MSELoss vs. WeightedMSELoss for Arousal/Valence regression. In what follows, we detail the workings of UnbalancedCrossEntropyLoss and WeightedMSELoss. We observe that UnbalancedCrossEntropyLoss presents a clear benefit over CrossEntropyLoss for the classification problem under consideration, while WeightedMSELoss typically does not manage to positively influence model performance for regression problems.

\subsubsection{UnbalancedCrossEntropyLoss}
As stated in the main text, UnbalancedCrossEntropyLoss ($\mathcal{L}_{\mbox{\tiny{UCE}}}$) allows to give different weights to different misclassifications. E.g., it allows to penalize classifying a ``joy'' as a ``sadness'' image heavier than classifying it as ``anticipation''. It is defined as
\begin{equation}
\mathcal{L}_{\mbox{\tiny{UCE}}} =
	\begin{cases}
		w_t \log{p_t} & t = h \\
		w_t \log{p_t} + w_{t,h} \log{1 - p_h} & t \neq h
	\end{cases} 
\end{equation}
with $t$ the target class with predicted probability $p_t$, $h$ the class with the highest predicted probability $p_h$, $w_t$ the weight of the target class, and $w_{t,h}$ the weight for misclassifying a sample of class $t$ as class $h$. In case $t=h$, this reverts to regular CrossEntropyLoss.

To be able to use UnbalancedCrossEntropy loss, a distance needs to be defined between each pair of output classes. For the Emo8 task, we use the shortest number of leaves between two emotions. E.g., the distance between ``joy'' and ``surprise'' is 3, and the distance between ``joy'' and ``anger'' is 2.

The class weight $w_{i}$ for class $i$ was computed according to
\begin{equation}
\label{eq:class_weight}
w_i = \frac{N}{N_c \cdot N_i},
\end{equation}
with $N$ the total number of samples, $N_c$ the number of classes and $N_i$ the number of samples of class $i$.

Finally, the weight $w_{i,j}$ for misclassifying a sample from class $i$ as class $j$ was computed as
\begin{equation}
w_{i,j} = \frac{d_{i,j}}{1 + w_{j}} \cdot w_{i},
\end{equation}
with $w_{i}$ the weight for class $i$, $w_{j}$ the weight for class $j$ and $d_{i,j}$ the distance between classes $i$ and $j$.

\subsubsection{WeightedMSELoss}
WeightedMSELoss ($\mathcal{L}_{\mbox{\tiny{WMSE}}}$) is a natural extension of the standard MSELoss to include class weights. Its mathematical formulation reads
\begin{equation}
\mathcal{L}_{\mbox{\tiny{WMSE}}} = \frac{1}{N} \sum_{i=1}^N \left( w_i \cdot \left(o_i - t_i\right)^2 \right),
\end{equation}
with $N$ the number of samples, $w_i$ the class weights as defined in Eq.\ref{eq:class_weight} and $o_i$ and $t_i$ the network output and target value for sample $i$ respectively.

\afterpage{
\begin{landscape}
\vspace*{\fill}
\begin{table}[!htbp]
	\footnotesize
	\centering
	\caption{Loss comparison for Emo8 classification and Arousal/Valence regression tasks, comparing test results for baseline models. Performance metrics format: $.xxx^{yy}$ should be read as average = $0.xxx$ (or $z.xxx$ if $z$ is specified), standard deviation = $.0yy$, taken over 10 runs. Best results in bold}
	\label{tb:baseline_cs_vs_uce}
	\begin{tabular}{@{}cl|l|ll|l|lllll|ll|l@{}}
& & \vtext{emonet} & \vtext{alexnet} & \vtext{alexnet p365} & \vtext{vgg16} & \vtext{resnet18} & \vtext{resnet18 p365} & \vtext{resnet50} & \vtext{resnet50 p365} & \vtext{resnet101} & \vtext{densenet161} & \vtext{densenet161 p365} & \vtext{CLIP} \\		
\midrule
		&&\multicolumn{12}{c}{\emph{Emo8}} \\
\multirow{3}{*}{\vtext{CE}} & Start LR & $0.0001$ & $0.0001$ & $0.0001$ & $0.001$ & $0.0001$ & $0.0001$ & $0.001$ & $0.0001$ & $0.001$ & $0.0001$ & $0.001$ & $0.001$ \\
		& Weighted F1 & $.178^{16}$ & $.273^{18}$ & $.237^{16}$ & $\mathbf{.311^{08}}$ & $.293^{04}$ & $.281^{07}$ & $.328^{13}$ & $.289^{09}$ & $\mathbf{.339^{09}}$ & $.316^{08}$ & $.298^{12}$ & $\mathbf{.440^{07}}$ \\
		& Avg.Prec. & $\mathbf{.148^{02}}$ & $\mathbf{.196^{06}}$ & $\mathbf{.162^{06}}$ & $\mathbf{.232^{05}}$ & $\mathbf{.223^{03}}$ & $\mathbf{.215^{06}}$ & $.249^{05}$ & $\mathbf{.228^{06}}$ & $\mathbf{.250^{03}}$ & $\mathbf{.247^{05}}$ & $\mathbf{.231^{02}}$ & $\mathbf{.346^{05}}$ \\
		\midrule
\multirow{3}{*}{\vtext{UCE}} & Start LR & $0.001$ & $0.0001$ & $0.0001$ & $0.001$ & $0.001$ & $0.001$ & $0.001$ & $0.001$ & $0.001$ & $0.001$ & $0.0001$ & $0.001$ \\
		& Weighted F1 & $\mathbf{.196^{18}}$ & $\mathbf{.276^{18}}$ & $\mathbf{.258^{14}}$ & $.310^{07}$ & $\mathbf{.298^{22}}$ & $\mathbf{.284^{12}}$ & $\mathbf{.335^{07}}$ & $\mathbf{.296^{13}}$ & $.335^{09}$ & $\mathbf{.327^{04}}$ & $\mathbf{.301^{09}}$ & $.431^{08}$ \\
		& Avg.Prec. & $.146^{02}$ & $.183^{10}$ & $.150^{07}$ & $.230^{03}$ & $.223^{04}$ & $.210^{03}$ & $\mathbf{.251^{04}}$ & $.227^{03}$ & $.247^{03}$ & $.244^{03}$ & $.228^{06}$ & $.340^{03}$ \\
		\multicolumn{14}{c}{\phantom{-}} \\
		&&\multicolumn{12}{c}{\emph{Arousal}} \\
\multirow{3}{*}{\vtext{MSE}} & Start LR & $0.001$ & $1e-05$ & $0.0001$ & $0.0001$ & $0.0001$ & $0.0001$ & $0.001$ & $0.0001$ & $0.001$ & $0.001$ & $0.0001$ & $0.001$ \\
		& MAE & $1.354^{01}$ & $\mathbf{1.333^{03}}$ & $\mathbf{1.344^{03}}$ & $\mathbf{1.320^{06}}$ & $\mathbf{1.334^{03}}$ & $\mathbf{1.333^{03}}$ & $\mathbf{1.311^{06}}$ & $\mathbf{1.327^{03}}$ & $\mathbf{1.314^{04}}$ & $1.320^{03}$ & $\mathbf{1.326^{04}}$ & $\mathbf{1.260^{06}}$ \\
		& Spearman R & $\mathbf{.094^{12}}$ & $\mathbf{.203^{10}}$ & $\mathbf{.173^{13}}$ & $\mathbf{.238^{14}}$ & $\mathbf{.214^{12}}$ & $.212^{07}$ & $\mathbf{.264^{10}}$ & $\mathbf{.226^{05}}$ & $.254^{12}$ & $\mathbf{.251^{10}}$ & $\mathbf{.231^{13}}$ & $\mathbf{.344^{10}}$ \\
		\midrule
\multirow{3}{*}{\vtext{WMSE}} & Start LR & $1e-05$ & $0.0001$ & $0.0001$ & $1e-05$ & $0.0001$ & $0.0001$ & $0.0001$ & $1e-05$ & $0.0001$ & $0.0001$ & $0.0001$ & $1e-05$ \\
		& MAE & $\mathbf{1.353^{02}}$ & $1.350^{07}$ & $1.361^{06}$ & $1.324^{06}$ & $1.338^{04}$ & $1.335^{05}$ & $1.318^{05}$ & $1.332^{04}$ & $1.317^{06}$ & $\mathbf{1.319^{05}}$ & $1.330^{04}$ & $1.265^{07}$ \\
		& Spearman R & $.072^{18}$ & $.193^{10}$ & $.171^{11}$ & $.238^{13}$ & $.208^{11}$ & $\mathbf{.213^{11}}$ & $.250^{09}$ & $.219^{10}$ & $\mathbf{.254^{11}}$ & $.249^{13}$ & $.227^{10}$ & $.340^{09}$ \\
		\multicolumn{14}{c}{\phantom{-}} \\
		&&\multicolumn{12}{c}{\emph{Valence}} \\
\multirow{3}{*}{\vtext{MSE}} & Start LR & $0.001$ & $1e-05$ & $0.0001$ & $0.001$ & $0.001$ & $0.001$ & $0.001$ & $0.001$ & $0.001$ & $0.001$ & $0.001$ & $0.0001$ \\
		& MAE & $\mathbf{1.516^{03}}$ & $\mathbf{1.363^{08}}$ & $\mathbf{1.398^{09}}$ & $\mathbf{1.312^{08}}$ & $\mathbf{1.385^{05}}$ & $\mathbf{1.382^{07}}$ & $\mathbf{1.290^{07}}$ & $\mathbf{1.344^{06}}$ & $\mathbf{1.276^{06}}$ & $\mathbf{1.336^{11}}$ & $\mathbf{1.355^{06}}$ & $\mathbf{1.015^{07}}$ \\
		& Spearman R & $.117^{12}$ & $.383^{10}$ & $.342^{10}$ & $\mathbf{.434^{09}}$ & $.394^{13}$ & $\mathbf{.378^{11}}$ & $\mathbf{.477^{07}}$ & $\mathbf{.414^{09}}$ & $\mathbf{.487^{08}}$ & $\mathbf{.440^{14}}$ & $.408^{10}$ & $\mathbf{.685^{05}}$ \\
		\midrule
\multirow{3}{*}{\vtext{WMSE}} & Start LR & $0.001$ & $1e-05$ & $0.0001$ & $0.001$ & $0.001$ & $0.001$ & $0.001$ & $0.001$ & $0.001$ & $0.001$ & $0.0001$ & $0.001$ \\
		& MAE & $1.564^{02}$ & $1.393^{09}$ & $1.424^{10}$ & $1.341^{11}$ & $1.412^{09}$ & $1.409^{10}$ & $1.310^{12}$ & $1.370^{09}$ & $1.295^{08}$ & $1.353^{07}$ & $1.378^{08}$ & $1.033^{09}$ \\
		& Spearman R & $\mathbf{.118^{12}}$ & $\mathbf{.386^{12}}$ & $\mathbf{.347^{13}}$ & $.432^{09}$ & $\mathbf{.395^{09}}$ & $.374^{09}$ & $.471^{14}$ & $.413^{11}$ & $.483^{09}$ & $.439^{07}$ & $\mathbf{.410^{10}}$ & $.676^{06}$ \\
	\end{tabular}
\end{table}
\vfill
\end{landscape}
}

\subsection{Additional Baseline Results}
\label{a:additional_baseline}
\begin{figure*}[htb]
\begin{center}
\centerline{\includegraphics[width=17cm]{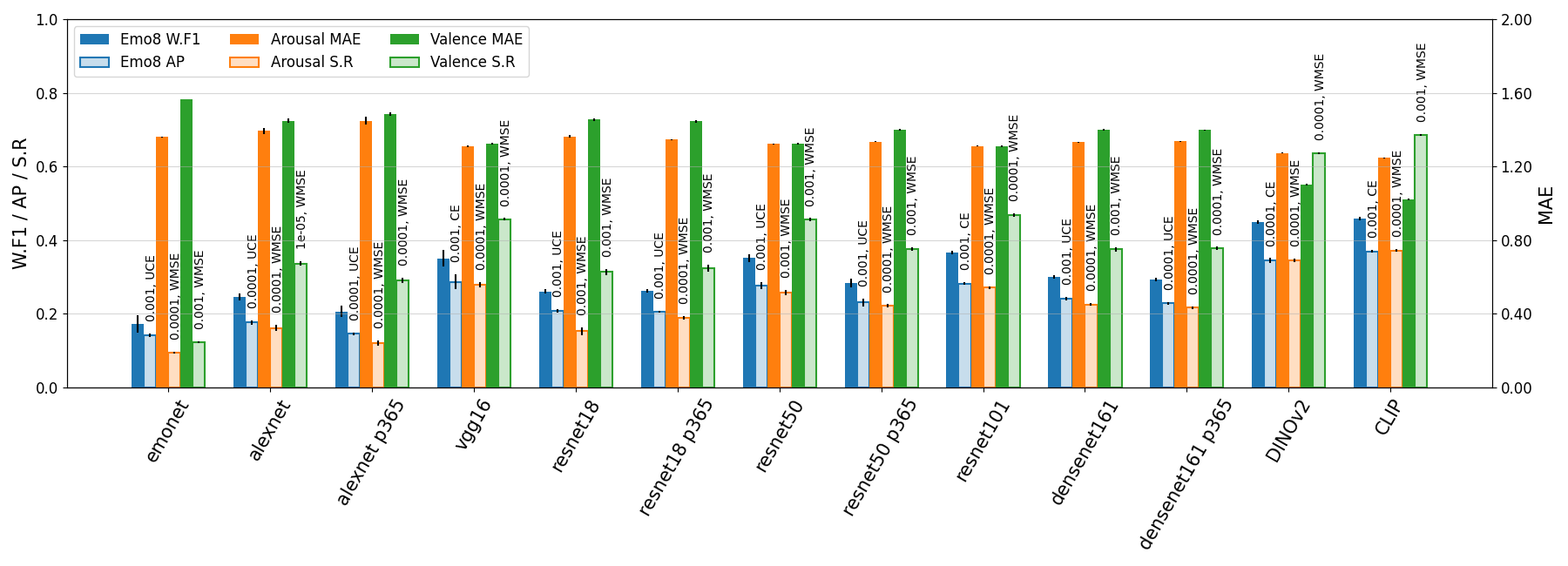}}
\caption{Train data baseline classification performance on the Emo8 classification and Arousal/Valence regression tasks. Metrics are: Weighted F1 (W.F1) and Average Precision (AP) for classification, and Mean Average Error (MAE) and Spearman R (S.R) for regression. The starting learning rate and loss corresponding to each model are displayed above the training bars. (U)CE = (Unbalanced)CrossEntropyLoss, (W)MSE = (Weighted)MeanSquaredErrorLoss, p365 = original model trained on Places365 dataset.}
\label{fig:barchart_baseline_train}
\end{center}
\end{figure*}

Baseline results for train data are depicted in \figref{fig:barchart_baseline_train}. Numerical baseline results on the train and test sets have been grouped in Tables~\ref{tb:baseline_models_train} and \ref{tb:baseline_models_test} respectively. A breakdown per Emo8 for train and test sets is shown in Tables~\ref{tb:baseline_perf_per_class_train} and \ref{tb:baseline_perf_per_class_test} respectively. For per class results, no Average Precision scores are reported, as we did not collect these.

\afterpage{
\begin{landscape}
\vspace*{\fill}
\begin{table}[!htbp]
	\footnotesize
	\centering
	\caption{Training results: Emo8 classification and Arousal/Valence regression performance for baseline models. Performance metrics format: $.xxx^{yy}$ should be read as average = $0.xxx$ (or $z.xxx$ if $z$ is specified), standard deviation = $.0yy$, taken over 10 runs.}
	\label{tb:baseline_models_train}
	\begin{tabular}{@{}l|l|ll|l|lllll|ll|ll@{}}
 & \vtext{emonet} & \vtext{alexnet} & \vtext{alexnet p365} & \vtext{vgg16} & \vtext{resnet18} & \vtext{resnet18 p365} & \vtext{resnet50} & \vtext{resnet50 p365} & \vtext{resnet101} & \vtext{densenet161} & \vtext{densenet161 p365} & \vtext{DINOv2} & \vtext{CLIP} \\		\midrule
		&\multicolumn{13}{c}{\emph{Emo8}} \\
		Loss & UCE & UCE & UCE & CE & UCE & UCE & UCE & UCE & CE & UCE & UCE & CE & CE \\
		Start LR & $0.001$ & $0.0001$ & $0.0001$ & $0.001$ & $0.001$ & $0.001$ & $0.001$ & $0.001$ & $0.001$ & $0.001$ & $0.0001$ & $0.0001$ & $0.001$ \\
		Accuracy & $.163^{22}$ & $.231^{10}$ & $.191^{15}$ & $.340^{24}$ & $.248^{05}$ & $.247^{05}$ & $.339^{11}$ & $.268^{12}$ & $.355^{05}$ & $.287^{05}$ & $.281^{04}$ & $.478^{04}$ & $.445^{04}$ \\
		F1 & $.144^{17}$ & $.211^{09}$ & $.175^{14}$ & $.315^{26}$ & $.222^{05}$ & $.223^{04}$ & $.308^{13}$ & $.244^{12}$ & $.321^{05}$ & $.258^{05}$ & $.249^{04}$ & $.345^{05}$ & $.392^{06}$ \\
		Weighted F1 & $.172^{24}$ & $.245^{10}$ & $.206^{15}$ & $.350^{23}$ & $.261^{06}$ & $.263^{04}$ & $.351^{12}$ & $.284^{12}$ & $.366^{05}$ & $.301^{05}$ & $.293^{04}$ & $.449^{05}$ & $.459^{05}$ \\
		Avg.Prec. & $.142^{05}$ & $.176^{07}$ & $.145^{04}$ & $.287^{21}$ & $.208^{05}$ & $.206^{03}$ & $.277^{10}$ & $.230^{10}$ & $.282^{03}$ & $.240^{05}$ & $.228^{04}$ & $.344^{07}$ & $.370^{05}$ \\
		\multicolumn{14}{c}{\phantom{-}} \\
		&\multicolumn{13}{c}{\emph{Arousal}} \\
		Loss & WMSE & MSE & MSE & MSE & MSE & MSE & MSE & MSE & MSE & WMSE & MSE & MSE & MSE \\
		Start LR & $1e-05$ & $1e-05$ & $0.0001$ & $0.0001$ & $0.0001$ & $0.0001$ & $0.001$ & $0.0001$ & $0.001$ & $0.0001$ & $0.0001$ & $0.0001$ & $0.001$ \\
		MAE & $1.352^{00}$ & $1.393^{42}$ & $1.387^{22}$ & $1.299^{03}$ & $1.341^{01}$ & $1.335^{02}$ & $1.310^{02}$ & $1.326^{02}$ & $1.309^{03}$ & $1.331^{02}$ & $1.328^{02}$ & $1.257^{03}$ & $1.251^{05}$ \\
		Spearman R & $0.071^{05}$ & $0.130^{26}$ & $0.115^{11}$ & $0.288^{07}$ & $0.177^{06}$ & $0.196^{05}$ & $0.259^{06}$ & $0.224^{05}$ & $0.260^{05}$ & $0.225^{03}$ & $0.216^{05}$ & $0.354^{06}$ & $0.361^{08}$ \\
		\multicolumn{14}{c}{\phantom{-}} \\
		&\multicolumn{13}{c}{\emph{Valence}} \\
		Loss & MSE & MSE & MSE & MSE & MSE & MSE & MSE & MSE & MSE & MSE & MSE & MSE & MSE \\
		Start LR & $0.001$ & $1e-05$ & $0.0001$ & $0.001$ & $0.001$ & $0.001$ & $0.001$ & $0.001$ & $0.001$ & $0.001$ & $0.001$ & $0.001$ & $0.0001$ \\
		MAE & $1.515^{02}$ & $1.435^{51}$ & $1.440^{15}$ & $1.315^{14}$ & $1.412^{03}$ & $1.404^{03}$ & $1.298^{05}$ & $1.368^{05}$ & $1.287^{04}$ & $1.368^{07}$ & $1.381^{04}$ & $1.098^{14}$ & $1.010^{02}$ \\
		Spearman R & $0.121^{03}$ & $0.313^{30}$ & $0.284^{12}$ & $0.431^{18}$ & $0.313^{06}$ & $0.326^{05}$ & $0.455^{05}$ & $0.369^{07}$ & $0.465^{05}$ & $0.371^{11}$ & $0.355^{05}$ & $0.627^{12}$ & $0.689^{02}$ \\
	\end{tabular}
\end{table}
\vfill
\end{landscape}
}

\afterpage{
\begin{landscape}
\vspace*{\fill}
\begin{table}[!htbp]
	\footnotesize
	\centering
	\caption{Test results: Emo8 classification and Arousal/Valence regression performance for baseline models. Performance metrics format: $.xxx^{yy}$ should be read as average = $0.xxx$ (or $z.xxx$ if $z$ is specified), standard deviation = $.0yy$, taken over 10 runs.}
	\label{tb:baseline_models_test}
	\begin{tabular}{@{}l|l|ll|l|lllll|ll|ll@{}}
 & \vtext{emonet} & \vtext{alexnet} & \vtext{alexnet p365} & \vtext{vgg16} & \vtext{resnet18} & \vtext{resnet18 p365} & \vtext{resnet50} & \vtext{resnet50 p365} & \vtext{resnet101} & \vtext{densenet161} & \vtext{densenet161 p365} & \vtext{DINOv2} & \vtext{CLIP} \\		\midrule
		&\multicolumn{13}{c}{\emph{Emo8}} \\
		Loss & UCE & UCE & UCE & CE & UCE & UCE & UCE & UCE & CE & UCE & UCE & CE & CE \\
		Start LR & $0.001$ & $0.0001$ & $0.0001$ & $0.001$ & $0.001$ & $0.001$ & $0.001$ & $0.001$ & $0.001$ & $0.001$ & $0.0001$ & $0.0001$ & $0.001$ \\
		Accuracy & $.192^{22}$ & $.280^{21}$ & $.259^{22}$ & $.303^{10}$ & $.301^{19}$ & $.286^{16}$ & $.327^{07}$ & $.296^{13}$ & $.329^{11}$ & $.330^{13}$ & $.294^{10}$ & $.450^{07}$ & $.427^{07}$ \\
		F1 & $.152^{10}$ & $.219^{09}$ & $.201^{09}$ & $.251^{07}$ & $.240^{14}$ & $.226^{07}$ & $.275^{05}$ & $.235^{11}$ & $.277^{06}$ & $.260^{05}$ & $.245^{08}$ & $.320^{08}$ & $.365^{07}$ \\
		Weighted F1 & $.196^{18}$ & $.276^{18}$ & $.258^{14}$ & $.311^{08}$ & $.298^{22}$ & $.284^{12}$ & $.335^{07}$ & $.296^{13}$ & $.339^{09}$ & $.327^{04}$ & $.301^{09}$ & $.424^{07}$ & $.440^{07}$ \\
		Avg.Prec. & $.146^{02}$ & $.183^{10}$ & $.150^{07}$ & $.232^{05}$ & $.223^{04}$ & $.210^{03}$ & $.251^{04}$ & $.227^{03}$ & $.250^{03}$ & $.244^{03}$ & $.228^{06}$ & $.310^{05}$ & $.346^{05}$ \\
		\multicolumn{14}{c}{\phantom{-}} \\
		&\multicolumn{13}{c}{\emph{Arousal}} \\
		Loss & WMSE & MSE & MSE & MSE & MSE & MSE & MSE & MSE & MSE & WMSE & MSE & MSE & MSE \\
		Start LR & $1e-05$ & $1e-05$ & $0.0001$ & $0.0001$ & $0.0001$ & $0.0001$ & $0.001$ & $0.0001$ & $0.001$ & $0.0001$ & $0.0001$ & $0.0001$ & $0.001$ \\
		MAE & $1.353^{02}$ & $1.333^{03}$ & $1.344^{03}$ & $1.320^{06}$ & $1.334^{03}$ & $1.333^{03}$ & $1.311^{06}$ & $1.327^{03}$ & $1.314^{04}$ & $1.319^{05}$ & $1.326^{04}$ & $1.276^{07}$ & $1.260^{06}$ \\
		Spearman R & $0.072^{18}$ & $0.203^{10}$ & $0.173^{13}$ & $0.238^{14}$ & $0.214^{12}$ & $0.212^{07}$ & $0.264^{10}$ & $0.226^{05}$ & $0.254^{12}$ & $0.249^{13}$ & $0.231^{13}$ & $0.322^{12}$ & $0.344^{10}$ \\
		\multicolumn{14}{c}{\phantom{-}} \\
		&\multicolumn{13}{c}{\emph{Valence}} \\
		Loss & MSE & MSE & MSE & MSE & MSE & MSE & MSE & MSE & MSE & MSE & MSE & MSE & MSE \\
		Start LR & $0.001$ & $1e-05$ & $0.0001$ & $0.001$ & $0.001$ & $0.001$ & $0.001$ & $0.001$ & $0.001$ & $0.001$ & $0.001$ & $0.001$ & $0.0001$ \\
		MAE & $1.516^{03}$ & $1.363^{08}$ & $1.398^{09}$ & $1.312^{08}$ & $1.385^{05}$ & $1.382^{07}$ & $1.290^{07}$ & $1.344^{06}$ & $1.276^{06}$ & $1.336^{11}$ & $1.355^{06}$ & $1.107^{12}$ & $1.015^{07}$ \\
		Spearman R & $0.117^{12}$ & $0.383^{10}$ & $0.342^{10}$ & $0.434^{09}$ & $0.394^{13}$ & $0.378^{11}$ & $0.477^{07}$ & $0.414^{09}$ & $0.487^{08}$ & $0.440^{14}$ & $0.408^{10}$ & $0.618^{09}$ & $0.685^{05}$ \\
	\end{tabular}
\end{table}
\vfill
\end{landscape}
}

\afterpage{
\begin{table}[!htbp]
	\footnotesize
	\centering
	\caption{Training results: Emo8 Recall, Precision and F1 metrics per emotion leaf for baseline models.  Format: $0.xx^{.yy}$ with $0.xx$ = average, $0.yy$ = standard deviation over 10 runs.}
	\label{tb:baseline_perf_per_class_train}
	\begin{tabular}{@{}l|llllllll@{}}
 & Joy & Trust & Fear & Surprise & Sadness & Disgust & Anger & Anticipation \\
		\midrule
		& \multicolumn{8}{c}{\emph{alexnet}} \\
		Recall & $0.25^{.01}$ & $0.20^{.01}$ & $0.25^{.01}$ & $0.19^{.02}$ & $0.30^{.02}$ & $0.20^{.01}$ & $0.32^{.02}$ & $0.16^{.01}$ \\
		Precision & $0.46^{.02}$ & $0.21^{.01}$ & $0.20^{.01}$ & $0.05^{.01}$ & $0.23^{.01}$ & $0.07^{.01}$ & $0.23^{.01}$ & $0.33^{.02}$ \\
		F1 & $0.33^{.02}$ & $0.20^{.01}$ & $0.22^{.01}$ & $0.08^{.01}$ & $0.26^{.02}$ & $0.10^{.01}$ & $0.27^{.01}$ & $0.21^{.01}$ \\
		& \multicolumn{8}{c}{\emph{alexnet p365}} \\
		Recall & $0.21^{.02}$ & $0.17^{.01}$ & $0.21^{.02}$ & $0.18^{.02}$ & $0.23^{.03}$ & $0.19^{.03}$ & $0.23^{.03}$ & $0.15^{.00}$ \\
		Precision & $0.41^{.03}$ & $0.18^{.01}$ & $0.16^{.01}$ & $0.05^{.01}$ & $0.19^{.02}$ & $0.06^{.01}$ & $0.18^{.02}$ & $0.30^{.03}$ \\
		F1 & $0.27^{.02}$ & $0.18^{.01}$ & $0.18^{.02}$ & $0.08^{.01}$ & $0.21^{.03}$ & $0.09^{.01}$ & $0.20^{.02}$ & $0.20^{.01}$ \\
		\midrule
		& \multicolumn{8}{c}{\emph{vgg16}} \\
		Recall & $0.36^{.02}$ & $0.28^{.03}$ & $0.36^{.02}$ & $0.36^{.07}$ & $0.45^{.03}$ & $0.35^{.06}$ & $0.47^{.02}$ & $0.23^{.02}$ \\
		Precision & $0.56^{.02}$ & $0.30^{.02}$ & $0.29^{.02}$ & $0.12^{.02}$ & $0.34^{.02}$ & $0.14^{.03}$ & $0.33^{.02}$ & $0.47^{.03}$ \\
		F1 & $0.44^{.02}$ & $0.29^{.03}$ & $0.32^{.02}$ & $0.18^{.04}$ & $0.39^{.03}$ & $0.20^{.04}$ & $0.39^{.02}$ & $0.31^{.02}$ \\
		\midrule
		& \multicolumn{8}{c}{\emph{resnet18}} \\
		Recall & $0.29^{.00}$ & $0.20^{.01}$ & $0.27^{.01}$ & $0.16^{.02}$ & $0.32^{.01}$ & $0.21^{.02}$ & $0.36^{.01}$ & $0.15^{.01}$ \\
		Precision & $0.48^{.01}$ & $0.22^{.01}$ & $0.22^{.01}$ & $0.05^{.00}$ & $0.24^{.01}$ & $0.07^{.01}$ & $0.24^{.01}$ & $0.36^{.01}$ \\
		F1 & $0.36^{.01}$ & $0.21^{.01}$ & $0.24^{.01}$ & $0.08^{.01}$ & $0.27^{.01}$ & $0.11^{.01}$ & $0.29^{.01}$ & $0.22^{.01}$ \\
		& \multicolumn{8}{c}{\emph{resnet18 p365}} \\
		Recall & $0.29^{.01}$ & $0.19^{.01}$ & $0.25^{.01}$ & $0.18^{.01}$ & $0.31^{.01}$ & $0.24^{.02}$ & $0.35^{.01}$ & $0.17^{.01}$ \\
		Precision & $0.47^{.01}$ & $0.22^{.01}$ & $0.21^{.01}$ & $0.06^{.00}$ & $0.24^{.01}$ & $0.08^{.00}$ & $0.24^{.01}$ & $0.37^{.01}$ \\
		F1 & $0.36^{.01}$ & $0.21^{.01}$ & $0.23^{.01}$ & $0.09^{.01}$ & $0.27^{.01}$ & $0.12^{.01}$ & $0.29^{.01}$ & $0.23^{.01}$ \\
		& \multicolumn{8}{c}{\emph{resnet50}} \\
		Recall & $0.38^{.01}$ & $0.28^{.02}$ & $0.34^{.01}$ & $0.29^{.04}$ & $0.46^{.01}$ & $0.30^{.03}$ & $0.46^{.02}$ & $0.23^{.01}$ \\
		Precision & $0.57^{.01}$ & $0.29^{.01}$ & $0.30^{.01}$ & $0.11^{.01}$ & $0.34^{.02}$ & $0.12^{.01}$ & $0.33^{.01}$ & $0.46^{.01}$ \\
		F1 & $0.46^{.01}$ & $0.28^{.01}$ & $0.32^{.01}$ & $0.16^{.02}$ & $0.39^{.01}$ & $0.17^{.02}$ & $0.38^{.01}$ & $0.31^{.01}$ \\
		& \multicolumn{8}{c}{\emph{resnet50 p365}} \\
		Recall & $0.30^{.01}$ & $0.20^{.01}$ & $0.27^{.02}$ & $0.22^{.03}$ & $0.34^{.02}$ & $0.26^{.03}$ & $0.38^{.01}$ & $0.19^{.01}$ \\
		Precision & $0.50^{.01}$ & $0.23^{.01}$ & $0.23^{.01}$ & $0.07^{.01}$ & $0.27^{.01}$ & $0.09^{.01}$ & $0.27^{.01}$ & $0.39^{.02}$ \\
		F1 & $0.38^{.01}$ & $0.22^{.01}$ & $0.25^{.01}$ & $0.10^{.02}$ & $0.30^{.01}$ & $0.13^{.02}$ & $0.32^{.01}$ & $0.25^{.02}$ \\
		& \multicolumn{8}{c}{\emph{resnet101}} \\
		Recall & $0.40^{.00}$ & $0.28^{.01}$ & $0.37^{.01}$ & $0.28^{.02}$ & $0.49^{.01}$ & $0.31^{.02}$ & $0.47^{.01}$ & $0.25^{.01}$ \\
		Precision & $0.58^{.01}$ & $0.30^{.01}$ & $0.31^{.01}$ & $0.11^{.01}$ & $0.35^{.01}$ & $0.13^{.01}$ & $0.33^{.01}$ & $0.47^{.01}$ \\
		F1 & $0.47^{.01}$ & $0.29^{.01}$ & $0.33^{.01}$ & $0.16^{.01}$ & $0.41^{.01}$ & $0.19^{.01}$ & $0.39^{.01}$ & $0.33^{.01}$ \\
		\midrule
		& \multicolumn{8}{c}{\emph{densenet161}} \\
		Recall & $0.34^{.01}$ & $0.23^{.01}$ & $0.30^{.01}$ & $0.19^{.03}$ & $0.39^{.01}$ & $0.25^{.02}$ & $0.40^{.01}$ & $0.18^{.01}$ \\
		Precision & $0.52^{.01}$ & $0.24^{.01}$ & $0.25^{.01}$ & $0.07^{.01}$ & $0.29^{.01}$ & $0.08^{.01}$ & $0.29^{.00}$ & $0.40^{.01}$ \\
		F1 & $0.41^{.01}$ & $0.24^{.01}$ & $0.27^{.01}$ & $0.10^{.01}$ & $0.33^{.01}$ & $0.13^{.01}$ & $0.34^{.01}$ & $0.25^{.01}$ \\
		& \multicolumn{8}{c}{\emph{densenet161 p365}} \\
		Recall & $0.33^{.01}$ & $0.21^{.01}$ & $0.28^{.01}$ & $0.15^{.02}$ & $0.37^{.01}$ & $0.24^{.02}$ & $0.42^{.01}$ & $0.19^{.01}$ \\
		Precision & $0.50^{.01}$ & $0.24^{.01}$ & $0.23^{.01}$ & $0.07^{.01}$ & $0.27^{.00}$ & $0.09^{.00}$ & $0.26^{.01}$ & $0.40^{.01}$ \\
		F1 & $0.40^{.01}$ & $0.22^{.01}$ & $0.25^{.01}$ & $0.09^{.01}$ & $0.31^{.01}$ & $0.13^{.01}$ & $0.32^{.01}$ & $0.26^{.01}$ \\
		\midrule
		& \multicolumn{8}{c}{\emph{DINOv2}} \\
		Recall & $0.71^{.00}$ & $0.22^{.01}$ & $0.27^{.01}$ & $0.00^{.00}$ & $0.51^{.01}$ & $0.04^{.01}$ & $0.43^{.01}$ & $0.59^{.01}$ \\
		Precision & $0.56^{.00}$ & $0.41^{.01}$ & $0.37^{.01}$ & $0.27^{.12}$ & $0.53^{.00}$ & $0.32^{.04}$ & $0.47^{.01}$ & $0.42^{.00}$ \\
		F1 & $0.62^{.00}$ & $0.28^{.01}$ & $0.31^{.01}$ & $0.01^{.00}$ & $0.52^{.01}$ & $0.07^{.01}$ & $0.45^{.01}$ & $0.49^{.00}$ \\
		& \multicolumn{8}{c}{\emph{CLIP}} \\
		Recall & $0.57^{.01}$ & $0.37^{.01}$ & $0.35^{.01}$ & $0.25^{.03}$ & $0.56^{.01}$ & $0.34^{.02}$ & $0.56^{.01}$ & $0.33^{.01}$ \\
		Precision & $0.69^{.01}$ & $0.35^{.01}$ & $0.33^{.01}$ & $0.11^{.01}$ & $0.53^{.01}$ & $0.15^{.01}$ & $0.45^{.00}$ & $0.52^{.01}$ \\
		F1 & $0.62^{.00}$ & $0.36^{.01}$ & $0.34^{.01}$ & $0.15^{.01}$ & $0.55^{.01}$ & $0.21^{.01}$ & $0.50^{.01}$ & $0.40^{.01}$ \\
	\end{tabular}
\end{table}
}

\afterpage{
\begin{table}[!htbp]
	\footnotesize
	\centering
	\caption{Test results: Emo8 Recall, Precision and F1 metrics per emotion leaf for baseline models.  Format: $0.xx^{.yy}$ with $0.xx$ = average, $0.yy$ = standard deviation over 10 runs.}
	\label{tb:baseline_perf_per_class_test}
	\begin{tabular}{@{}l|llllllll@{}}
 & Joy & Trust & Fear & Surprise & Sadness & Disgust & Anger & Anticipation \\
		\midrule
		& \multicolumn{8}{c}{\emph{alexnet}} \\
		Recall & $0.38^{.09}$ & $0.18^{.10}$ & $0.24^{.08}$ & $0.05^{.05}$ & $0.36^{.09}$ & $0.08^{.05}$ & $0.38^{.08}$ & $0.23^{.08}$ \\
		Precision & $0.46^{.04}$ & $0.22^{.02}$ & $0.20^{.02}$ & $0.04^{.01}$ & $0.25^{.04}$ & $0.07^{.01}$ & $0.25^{.03}$ & $0.35^{.02}$ \\
		F1 & $0.40^{.06}$ & $0.18^{.05}$ & $0.21^{.03}$ & $0.04^{.02}$ & $0.29^{.01}$ & $0.07^{.02}$ & $0.29^{.02}$ & $0.27^{.05}$ \\
		& \multicolumn{8}{c}{\emph{alexnet p365}} \\
		Recall & $0.40^{.09}$ & $0.17^{.08}$ & $0.24^{.09}$ & $0.10^{.09}$ & $0.28^{.08}$ & $0.08^{.06}$ & $0.24^{.08}$ & $0.20^{.09}$ \\
		Precision & $0.42^{.02}$ & $0.22^{.02}$ & $0.19^{.02}$ & $0.05^{.01}$ & $0.24^{.04}$ & $0.05^{.02}$ & $0.23^{.03}$ & $0.33^{.03}$ \\
		F1 & $0.40^{.04}$ & $0.18^{.05}$ & $0.20^{.03}$ & $0.06^{.02}$ & $0.25^{.02}$ & $0.05^{.03}$ & $0.22^{.04}$ & $0.24^{.06}$ \\
		\midrule
		& \multicolumn{8}{c}{\emph{vgg16}} \\
		Recall & $0.40^{.04}$ & $0.20^{.03}$ & $0.26^{.05}$ & $0.09^{.04}$ & $0.38^{.04}$ & $0.14^{.05}$ & $0.40^{.08}$ & $0.25^{.07}$ \\
		Precision & $0.50^{.01}$ & $0.24^{.02}$ & $0.23^{.02}$ & $0.05^{.01}$ & $0.30^{.02}$ & $0.07^{.02}$ & $0.30^{.05}$ & $0.38^{.03}$ \\
		F1 & $0.44^{.02}$ & $0.21^{.02}$ & $0.24^{.02}$ & $0.06^{.02}$ & $0.33^{.02}$ & $0.09^{.02}$ & $0.34^{.02}$ & $0.29^{.04}$ \\
		\midrule
		& \multicolumn{8}{c}{\emph{resnet18}} \\
		Recall & $0.36^{.09}$ & $0.31^{.11}$ & $0.37^{.09}$ & $0.04^{.04}$ & $0.30^{.11}$ & $0.09^{.05}$ & $0.33^{.10}$ & $0.27^{.07}$ \\
		Precision & $0.50^{.05}$ & $0.22^{.02}$ & $0.23^{.02}$ & $0.04^{.03}$ & $0.31^{.07}$ & $0.10^{.02}$ & $0.30^{.03}$ & $0.36^{.04}$ \\
		F1 & $0.40^{.08}$ & $0.24^{.04}$ & $0.28^{.02}$ & $0.03^{.03}$ & $0.28^{.06}$ & $0.09^{.02}$ & $0.31^{.04}$ & $0.30^{.04}$ \\
		& \multicolumn{8}{c}{\emph{resnet18 p365}} \\
		Recall & $0.36^{.08}$ & $0.19^{.07}$ & $0.30^{.09}$ & $0.04^{.02}$ & $0.30^{.10}$ & $0.12^{.06}$ & $0.35^{.08}$ & $0.28^{.10}$ \\
		Precision & $0.47^{.03}$ & $0.24^{.02}$ & $0.19^{.02}$ & $0.06^{.03}$ & $0.27^{.04}$ & $0.08^{.02}$ & $0.25^{.03}$ & $0.35^{.03}$ \\
		F1 & $0.40^{.04}$ & $0.20^{.04}$ & $0.23^{.03}$ & $0.04^{.02}$ & $0.27^{.03}$ & $0.08^{.02}$ & $0.28^{.03}$ & $0.30^{.07}$ \\
		& \multicolumn{8}{c}{\emph{resnet50}} \\
		Recall & $0.39^{.05}$ & $0.25^{.04}$ & $0.32^{.06}$ & $0.07^{.02}$ & $0.44^{.06}$ & $0.13^{.02}$ & $0.44^{.04}$ & $0.28^{.03}$ \\
		Precision & $0.54^{.02}$ & $0.24^{.02}$ & $0.27^{.03}$ & $0.06^{.02}$ & $0.33^{.03}$ & $0.08^{.01}$ & $0.31^{.02}$ & $0.40^{.02}$ \\
		F1 & $0.45^{.03}$ & $0.24^{.02}$ & $0.29^{.02}$ & $0.06^{.01}$ & $0.37^{.02}$ & $0.10^{.01}$ & $0.36^{.01}$ & $0.33^{.02}$ \\
		& \multicolumn{8}{c}{\emph{resnet50 p365}} \\
		Recall & $0.38^{.08}$ & $0.19^{.11}$ & $0.31^{.10}$ & $0.06^{.06}$ & $0.30^{.07}$ & $0.10^{.05}$ & $0.33^{.09}$ & $0.30^{.11}$ \\
		Precision & $0.49^{.05}$ & $0.24^{.02}$ & $0.21^{.02}$ & $0.04^{.02}$ & $0.31^{.05}$ & $0.07^{.01}$ & $0.30^{.05}$ & $0.35^{.03}$ \\
		F1 & $0.42^{.04}$ & $0.19^{.06}$ & $0.24^{.03}$ & $0.04^{.03}$ & $0.30^{.02}$ & $0.08^{.02}$ & $0.30^{.03}$ & $0.31^{.05}$ \\
		& \multicolumn{8}{c}{\emph{resnet101}} \\
		Recall & $0.42^{.03}$ & $0.23^{.04}$ & $0.34^{.05}$ & $0.08^{.03}$ & $0.45^{.04}$ & $0.15^{.05}$ & $0.42^{.03}$ & $0.25^{.04}$ \\
		Precision & $0.54^{.01}$ & $0.25^{.01}$ & $0.26^{.02}$ & $0.05^{.01}$ & $0.33^{.02}$ & $0.08^{.01}$ & $0.31^{.01}$ & $0.43^{.03}$ \\
		F1 & $0.47^{.02}$ & $0.24^{.02}$ & $0.29^{.02}$ & $0.06^{.01}$ & $0.38^{.01}$ & $0.10^{.02}$ & $0.36^{.01}$ & $0.31^{.03}$ \\
		\midrule
		& \multicolumn{8}{c}{\emph{densenet161}} \\
		Recall & $0.48^{.07}$ & $0.17^{.06}$ & $0.26^{.07}$ & $0.06^{.05}$ & $0.39^{.09}$ & $0.13^{.09}$ & $0.43^{.05}$ & $0.27^{.05}$ \\
		Precision & $0.49^{.03}$ & $0.28^{.02}$ & $0.25^{.02}$ & $0.04^{.02}$ & $0.31^{.04}$ & $0.08^{.02}$ & $0.31^{.02}$ & $0.38^{.01}$ \\
		F1 & $0.48^{.02}$ & $0.20^{.04}$ & $0.25^{.04}$ & $0.04^{.03}$ & $0.34^{.02}$ & $0.09^{.02}$ & $0.36^{.01}$ & $0.31^{.04}$ \\
		& \multicolumn{8}{c}{\emph{densenet161 p365}} \\
		Recall & $0.38^{.03}$ & $0.20^{.03}$ & $0.27^{.04}$ & $0.05^{.03}$ & $0.38^{.02}$ & $0.15^{.02}$ & $0.43^{.03}$ & $0.22^{.03}$ \\
		Precision & $0.49^{.02}$ & $0.23^{.01}$ & $0.23^{.01}$ & $0.04^{.01}$ & $0.28^{.01}$ & $0.08^{.01}$ & $0.25^{.02}$ & $0.38^{.01}$ \\
		F1 & $0.43^{.02}$ & $0.21^{.02}$ & $0.25^{.02}$ & $0.04^{.02}$ & $0.32^{.02}$ & $0.10^{.01}$ & $0.32^{.01}$ & $0.28^{.03}$ \\
		\midrule
		& \multicolumn{8}{c}{\emph{DINOv2}} \\
		Recall & $0.67^{.05}$ & $0.23^{.05}$ & $0.25^{.07}$ & $0.00^{.00}$ & $0.47^{.02}$ & $0.02^{.02}$ & $0.42^{.05}$ & $0.55^{.04}$ \\
		Precision & $0.56^{.03}$ & $0.36^{.03}$ & $0.32^{.01}$ & $0.00^{.00}$ & $0.49^{.04}$ & $0.17^{.08}$ & $0.42^{.03}$ & $0.40^{.01}$ \\
		F1 & $0.61^{.01}$ & $0.27^{.04}$ & $0.28^{.04}$ & $0.00^{.00}$ & $0.48^{.02}$ & $0.04^{.03}$ & $0.42^{.02}$ & $0.46^{.01}$ \\
		& \multicolumn{8}{c}{\emph{CLIP}} \\
		Recall & $0.57^{.03}$ & $0.33^{.03}$ & $0.35^{.05}$ & $0.17^{.02}$ & $0.54^{.02}$ & $0.21^{.03}$ & $0.52^{.03}$ & $0.34^{.04}$ \\
		Precision & $0.67^{.01}$ & $0.33^{.01}$ & $0.32^{.02}$ & $0.08^{.01}$ & $0.52^{.03}$ & $0.11^{.01}$ & $0.43^{.02}$ & $0.48^{.01}$ \\
		F1 & $0.61^{.02}$ & $0.33^{.02}$ & $0.33^{.03}$ & $0.11^{.01}$ & $0.53^{.02}$ & $0.14^{.01}$ & $0.47^{.01}$ & $0.40^{.02}$ \\
	\end{tabular}
\end{table}
}

\subsection{Additional Beyond Baseline Results}
\label{a:additional_beyond_baseline}
\begin{figure}[htb]
\begin{center}
\centerline{\includegraphics[width=17cm]{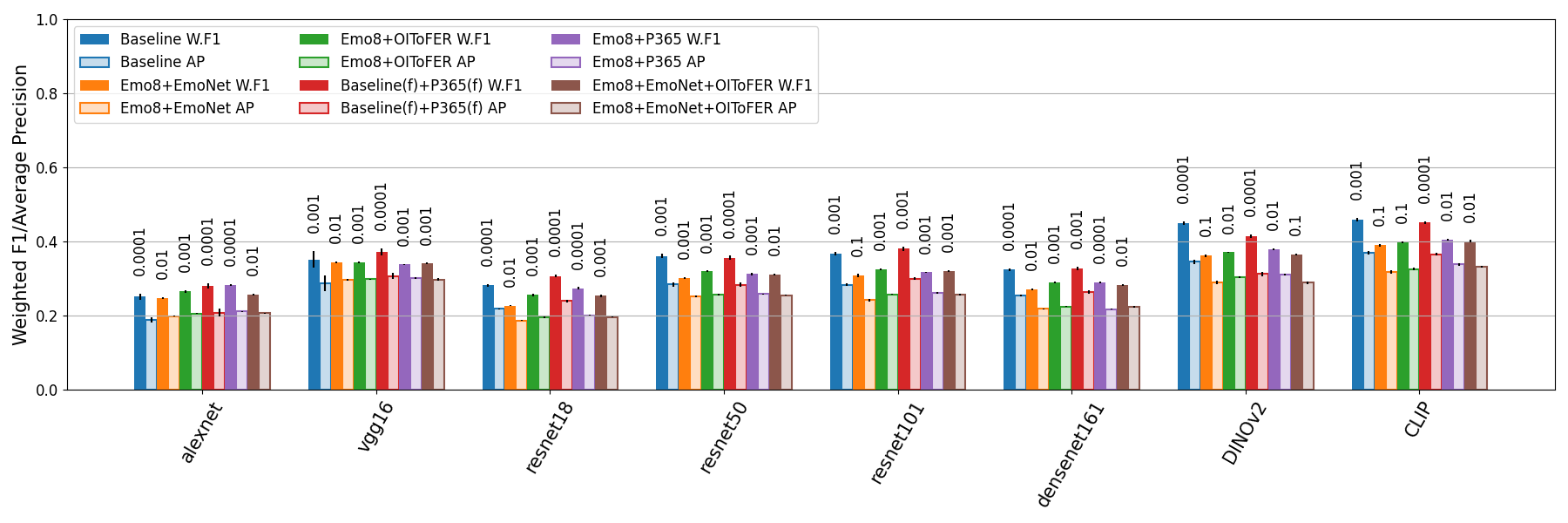}}
\caption{Training data results for extensions beyond the ImageNet baseline by applying late fusion with EmoNet predictions (EmoNet), Facial Emotion Recognition predictions (OIToFER) and Places365 (P365) predictions or features. For all models, predictions on the dataset are concatenated and sent through a linear layer, except when `(f)' is shown, indicating model features are concatenated. The starting learning rate corresponding to each model is displayed above the training bars.}
\label{fig:beyond_baseline_emo8_train}
\end{center}
\end{figure}

\begin{figure}[!htb]
\begin{center}
\centerline{\includegraphics[width=17cm]{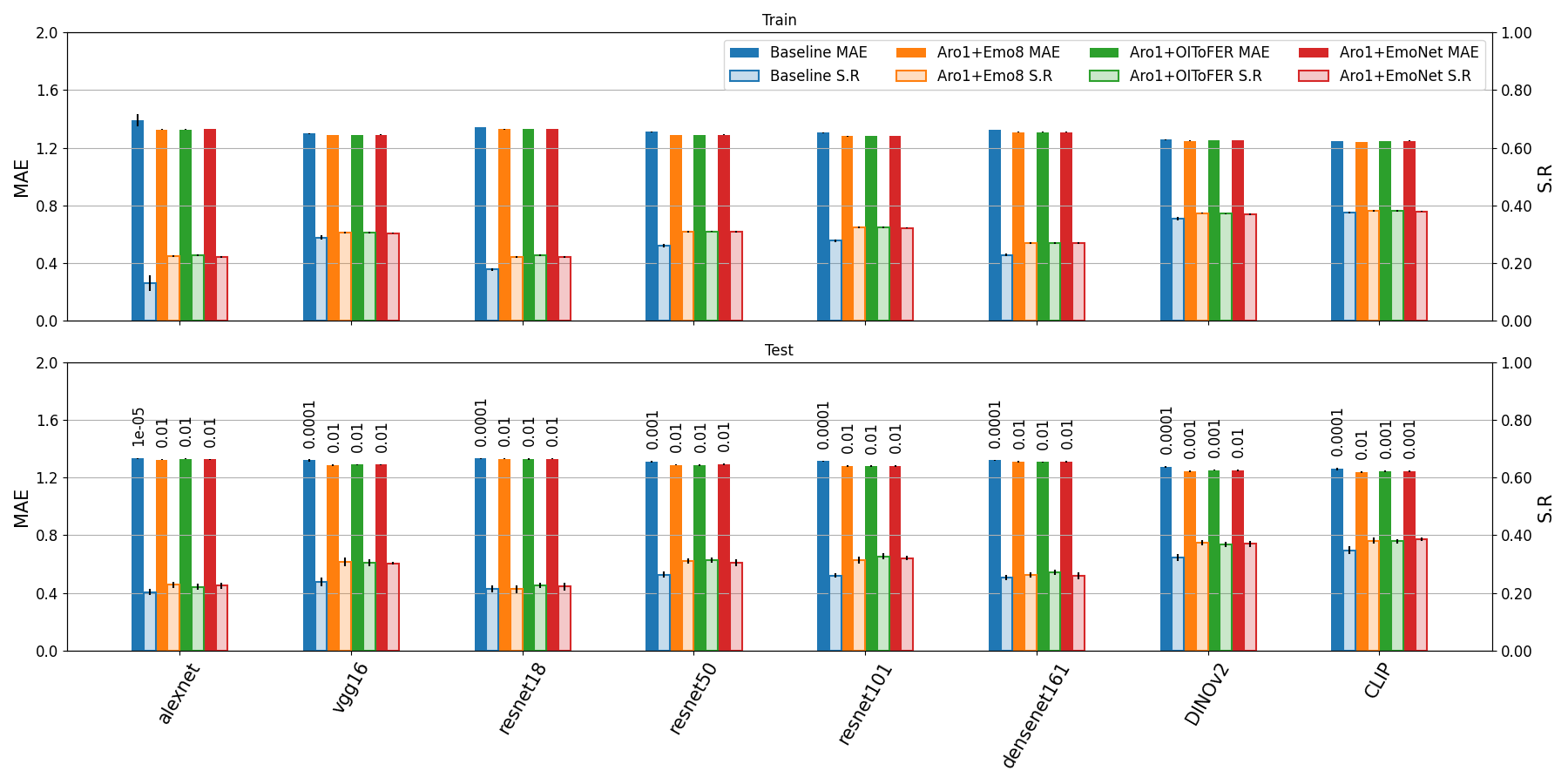}}
\caption{Arousal regression results for extensions beyond the baseline by applying late fusion with precomputed Emo8 predictions of the same architecture (Emo8), Facial Emotion Recognition predictions (OIToFER) and EmoNet predictions (EmoNet). For all models, precomputed predictions on the dataset (Aro1) are concatenated and sent through a linear layer. Metrics are: Mean Average Error (MAE) and Spearman R (S.R). The starting learning rate corresponding to each model is displayed above the training bars.}
\label{fig:beyond_baseline_aro1}
\end{center}
\end{figure}

\begin{figure}[!htb]
\begin{center}
\centerline{\includegraphics[width=17cm]{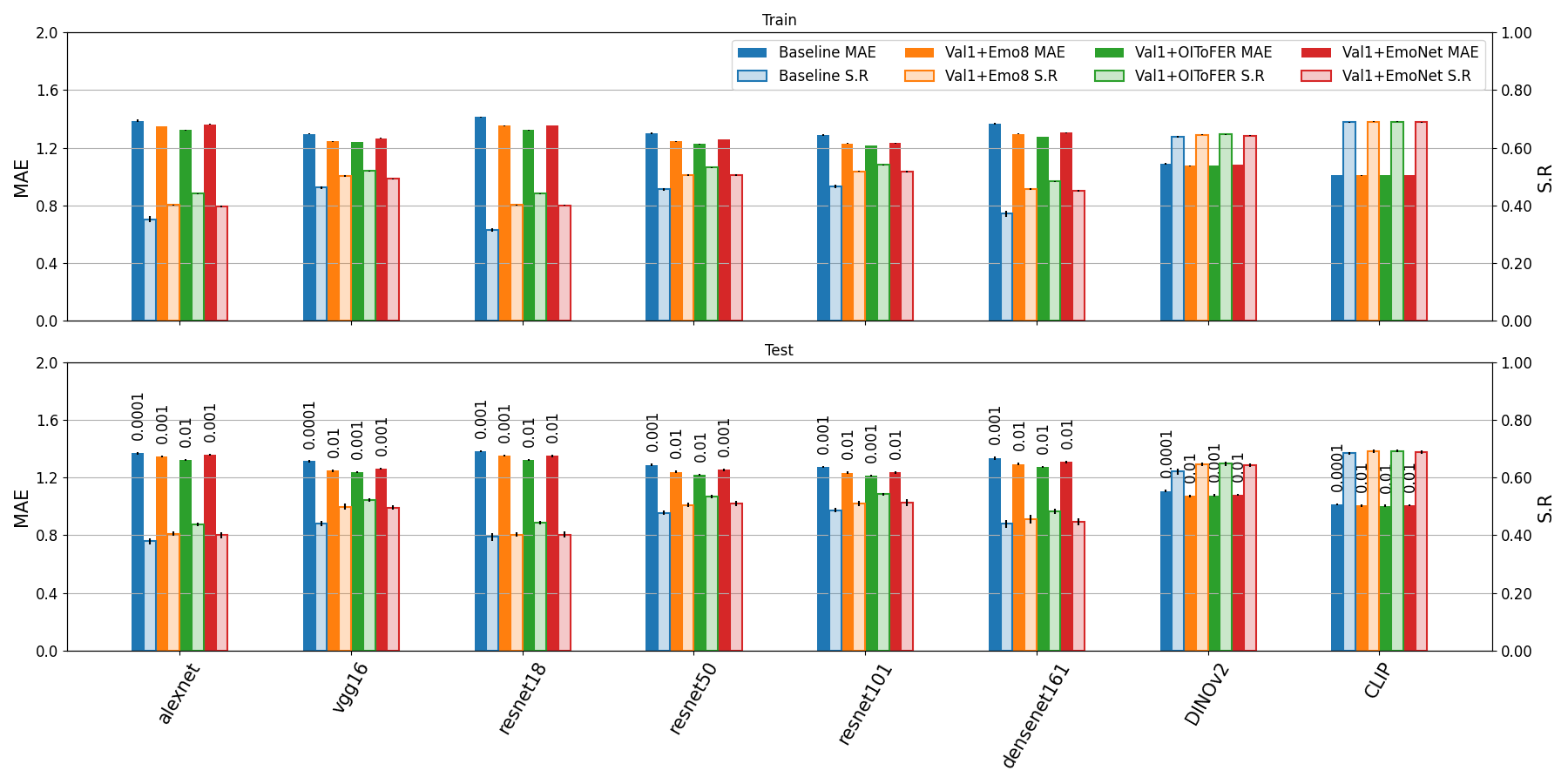}}
\caption{Valence regression results for extensions beyond the baseline by applying late fusion with precomputed Emo8 predictions of the same architecture (Emo8), Facial Emotion Recognition predictions (OIToFER) and EmoNet predictions (EmoNet). For all models, precomputed predictions on the dataset (Val1) are concatenated and sent through a linear layer. Metrics are: Mean Average Error (MAE) and Spearman R (S.R). The starting learning rate corresponding to each model is displayed above the training bars.}
\label{fig:beyond_baseline_val1}
\end{center}
\end{figure}

The beyond baseline Emo8 classification results on train data are shown in \figref{fig:beyond_baseline_emo8_train}. Barcharts depicting the beyond baseline results for the Arousal and Valence regression tasks are grouped in Figures~\ref{fig:beyond_baseline_aro1} and \ref{fig:beyond_baseline_val1} respectively.

For the Arousal and Valence regression tasks, we dropped experiments with Places365 models in favor of precomputed Emo8 predictions using the same model architecture. E.g., for the Arousal task and AlexNet architecture, we combine the 8-feature vector obtained by applying a pretrained AlexNet Emo8 classifier with the 1-feature vector obtained by applying a pretrained AlexNet Arousal regressor. We also changed the merger network from simply concatenating the stream features to first reducing the second stream features to 1D by means of a linear layer + sigmoid, and then concatening both 1D features (from the precomputed Arousal/Valence  regression + reduced second stream) and send these through a final linear layer + sigmoid. Finally, we only consider MSELoss, and $\mbox{lr}_0 \in [10^{-3}, 10^{-4}, 10^{-5}]$.

A numerical comparison of the baseline to the (overall best performing) ``Baseline+OIToFER'' model for all three tasks is included in \tabref{tb:baseline_vs_oitofer_aro_val}. From this, it is apparent that obtainable gains are architecture-dependent, with the VGG16 and ResNet50 architectures obtaining most gains and the DenseNet161, CLIP and DINOv2 architectures barely improving, regardless of task. Obtained gains are highest for the Emo8 task (absolute 13.7\% AP gain, or relative 59\%, for VGG16). Gains for the Arousal and Valence tasks are somewhat less straightforward to compare, as changes in MAE and Spearman R do not always agree. Compare, e.g., for ResNet50, for the Valence task an absolute 0.121 MAE and 0.120 Spearman R improvement, or relative 9.0\% and 22.5\% respectively, to an absolute 0.039 MAE and 0.087 Spearman R improvement, or relative 2.9\% and 38\% respectively for the Arousal task.

\begin{table}[!htbp]
	\footnotesize
	\centering
	\caption{Baseline vs. +OIToFER: A comparison of Emo8 classification and Arousal/Valence regression performance on the test data.}
	\label{tb:baseline_vs_oitofer_aro_val}
	\begin{tabular}{@{}cl|l|l|lll|l|ll@{}}
& & \vtext{alexnet} & \vtext{vgg16} & \vtext{resnet18} & \vtext{resnet50} & \vtext{resnet101} & \vtext{densenet161} & \vtext{DINOv2} & \vtext{CLIP} \\
		\midrule
		&&\multicolumn{8}{c}{\emph{Emo8}} \\
\multirow{5}{*}{\vtext{Baseline}} & Start LR & $0.0001$ & $0.001$ & $0.0001$ & $0.0001$ & $0.001$ & $0.0001$ & $0.0001$ & $0.001$ \\
		& Accuracy & $.271^{25}$ & $.303^{10}$ & $.271^{08}$ & $.279^{09}$ & $.329^{11}$ & $.308^{09}$ & $.450^{07}$ & $.427^{07}$ \\
		& F1 & $.221^{10}$ & $.251^{07}$ & $.232^{06}$ & $.240^{08}$ & $.277^{06}$ & $.261^{08}$ & $.320^{08}$ & $.365^{07}$ \\
		& Weighted F1 & $.273^{18}$ & $.311^{08}$ & $.281^{07}$ & $.289^{09}$ & $.339^{09}$ & $.316^{08}$ & $.424^{07}$ & $.440^{07}$ \\
		& Avg.Prec. & $.196^{06}$ & $.232^{05}$ & $.215^{06}$ & $.228^{06}$ & $.250^{03}$ & $.247^{05}$ & $.310^{05}$ & $.346^{05}$ \\
		\midrule
\multirow{5}{*}{\vtext{+OIToFER}} & Start LR & $0.001$ & $0.001$ & $0.001$ & $0.001$ & $0.001$ & $0.001$ & $0.01$ & $0.1$ \\
		& Accuracy & $.299^{04}$ & $.390^{05}$ & $.315^{06}$ & $.366^{05}$ & $.367^{07}$ & $.343^{04}$ & $.451^{03}$ & $.475^{10}$ \\
		& F1 & $.243^{03}$ & $.363^{05}$ & $.239^{06}$ & $.321^{04}$ & $.321^{07}$ & $.278^{05}$ & $.288^{05}$ & $.322^{16}$ \\
		& Weighted F1 & $.296^{04}$ & $.399^{05}$ & $.299^{07}$ & $.372^{04}$ & $.372^{07}$ & $.339^{04}$ & $.395^{05}$ & $.430^{20}$ \\
		& Avg.Prec. & $.237^{05}$ & $.369^{05}$ & $.232^{02}$ & $.311^{05}$ & $.313^{06}$ & $.269^{03}$ & $.360^{07}$ & $.384^{06}$ \\
		\multicolumn{10}{c}{\phantom{-}} \\
		&&\multicolumn{8}{c}{\emph{Arousal}} \\
\multirow{3}{*}{\vtext{Baseline}} & Start LR & $1e-05$ & $0.0001$ & $0.0001$ & $0.0001$ & $0.0001$ & $0.0001$ & $0.0001$ & $0.0001$ \\
		& MAE & $1.333^{03}$ & $1.320^{06}$ & $1.333^{03}$ & $1.327^{03}$ & $1.314^{04}$ & $1.321^{03}$ & $1.276^{07}$ & $1.260^{07}$ \\
		& Spearman R & $.203^{10}$ & $.238^{14}$ & $.212^{07}$ & $.226^{05}$ & $.261^{08}$ & $.253^{08}$ & $.322^{12}$ & $.347^{14}$ \\
		\midrule
\multirow{3}{*}{\vtext{+OIToFER}} & Start LR & $0.01$ & $0.01$ & $0.01$ & $0.01$ & $0.01$ & $0.01$ & $0.001$ & $0.001$ \\
		& MAE & $1.330^{03}$ & $1.290^{05}$ & $1.329^{04}$ & $1.288^{05}$ & $1.279^{06}$ & $1.308^{04}$ & $1.253^{04}$ & $1.245^{05}$ \\
		& Spearman R & $.220^{11}$ & $.305^{12}$ & $.226^{10}$ & $.313^{09}$ & $.326^{11}$ & $.272^{09}$ & $.369^{09}$ & $.380^{07}$ \\
		\multicolumn{10}{c}{\phantom{-}} \\
		&&\multicolumn{8}{c}{\emph{Valence}} \\
\multirow{3}{*}{\vtext{Baseline}} & Start LR & $0.0001$ & $0.0001$ & $0.001$ & $0.001$ & $0.001$ & $0.001$ & $0.0001$ & $0.0001$ \\
		& MAE & $1.369^{08}$ & $1.314^{06}$ & $1.382^{07}$ & $1.344^{06}$ & $1.276^{06}$ & $1.336^{11}$ & $1.107^{09}$ & $1.015^{07}$ \\
		& Spearman R & $.379^{09}$ & $.440^{08}$ & $.378^{11}$ & $.414^{09}$ & $.487^{08}$ & $.440^{14}$ & $.621^{10}$ & $.685^{05}$ \\
		\midrule
\multirow{3}{*}{\vtext{+OIToFER}} & Start LR & $0.01$ & $0.001$ & $0.01$ & $0.01$ & $0.001$ & $0.01$ & $0.001$ & $0.01$ \\
		& MAE & $1.324^{06}$ & $1.237^{06}$ & $1.324^{07}$ & $1.223^{06}$ & $1.215^{07}$ & $1.274^{06}$ & $1.077^{08}$ & $1.005^{07}$ \\
		& Spearman R & $.438^{07}$ & $.522^{06}$ & $.444^{07}$ & $.534^{06}$ & $.543^{05}$ & $.483^{09}$ & $.649^{07}$ & $.693^{05}$ \\
	\end{tabular}
\end{table}

\subsection{A Note on the Fuzziness of Emotion Recognition}
\label{a:disc_single_anns}
As shown in \S\ref{a:inter_ann}, if there is disagreement between annotators concerning the emotion depicted in an image, then it is typically among similar emotions. So, although annotators often disagree, the different labels provided for a same image are far from random, instead showing clear tendencies toward a specific region of the emotion spectrum.

This fuzzyiness in assigned labels is a feature of human psychology. Emotion recognition is hard, nuanced, and multidimensional. With more raters, one would obtain a distribution of responses, but still no perfect agreement. To compound this issue, the estimate of the distribution per image would be poor unless one has many raters per image (tens of raters for tens of thousands of images!). This is, for many reasons not least of which financially, highly impractical.

Having one rater per image gives uncertainty if one is interested in a single image, but the average performance across many images is still meaningful. To demonstrate this, we perform the following experiment: for several architectures, we train an Emo8 prediction model on our dataset, and once trained, we let the model make predictions for each image in the training set. We train 5 models per $\mbox{lr}_0 \in [10^{-1}, 10^{-2}, 10^{-3}, 10^{-4}]$ using CrossEntropyLoss, and keep the one with the highest Weighted F1 score as the winner. For these models, we list in \tabref{tb:rank_x} how often the annotated emotion was ranked $N$ (out of 8), and in \tabref{tb:dist_x} we show the distance between the top predicted and annotated emotions. Except for AlexNet, all other models show a nice downward sloping behavior as either the rank (\tabref{tb:rank_x}) or distance (\tabref{tb:dist_x}) increases. In other words, the ``mistakes'' made by these models are clearly not random, but show behavior that is similar to those observed in the human annotators. This confirms that even with a single annotation per image, already valuable results and insights can be obtained.

\begin{table}[!htbp]
	\centering
	\caption{Percentage of times, with respect to the full training set, the annotated emotion was ranked $N$ in the model predictions.}
	\label{tb:rank_x}
	\begin{tabular}{@{}l|llllllll@{}}
	& 0 & 1 & 2 & 3 & 4 & 5 & 6 & 7 \\
	\midrule
	alexnet & 22.5 & 15.7 & 13.5 & 11.8 & 11.0 & 10.1 & 8.9 & 6.5 \\
	vgg16 & 41.8 & 20.3 & 13.2 & 9.0 & 6.0 & 4.7 & 3.1 & 2.0 \\
	resnet18 & 31.3 & 19.2 & 13.3 & 10.6 & 8.4 & 6.9 & 5.8 & 4.5 \\
	resnet50 & 38.1 & 19.4 & 13.5 & 9.5 & 7.0 & 5.5 & 4.0 & 2.9 \\
	resnet101 & 38.4 & 20.2 & 13.2 & 9.4 & 6.8 & 5.1 & 4.0 & 3.0 \\
	densenet161 & 33.3 & 18.7 & 13.4 & 10.4 & 8.4 & 6.5 & 5.2 & 4.0
	\end{tabular}
\end{table}

\begin{table}[!htbp]
	\centering
	\caption{Percentage of times, with respect to the full training set, the distance between the annotated and predicted emotions was equal to $N$ in the model predictions.}
	\label{tb:dist_x}
	\begin{tabular}{@{}l|lllll@{}}
	& 0 & 1 & 2 & 3 & 4 \\
	\midrule
	alexnet & 22.5 & 19.7 & 19.1 & 23.7 & 15.0 \\
	vgg16 & 41.8 & 21.4 & 15.2 & 14.0 & 7.7 \\
	resnet18 & 31.3 & 26.1 & 17.0 & 17.0 & 8.6 \\
	resnet50 & 38.1 & 21.7 & 16.2 & 15.7 & 8.3 \\
	resnet101 & 38.4 & 23.1 & 16.9 & 14.1 & 7.5 \\
	densenet161 & 33.3 & 22.8 & 18.4 & 16.8 & 8.7
	\end{tabular}
\end{table}

\subsection{Author Responsibility Statement}
We, the authors, confirm that we bear all responsibility in case of any violation of rights during the collection of the data or other work, and that we will take appropriate action if and when needed, e.g., to remove data with such issues. We also confirm the licenses provided with the data and code associated with this work: an MIT license for all code; a CC BY-NC-SA 4.0 license for the dataset (concretely, the list of URLs and the annotations).

In particular, and as clearly and explicitly stated on our repository (under ``Legal Compliance and Privacy''), we invite any rightful copyright holders or persons depicted in any of the images that do not want their work/likeness to be used within the context of this dataset to contact us, so that we can remove that specific material from the dataset.

\end{document}